\documentclass[conference]{IEEEtran}
\usepackage{times} 

\usepackage[sort&compress,numbers]{natbib}
\usepackage{multicol}
\usepackage[bookmarks=true]{hyperref}

\makeatother
\usepackage{amsfonts}

\usepackage{mathtools}
\usepackage{tabularx}
\usepackage{graphicx}
\usepackage{subfigure}
\usepackage{enumerate}
\usepackage{float}
\usepackage{url}
\usepackage{verbatim}
\usepackage{array}
\usepackage{multirow}
\usepackage{longtable}
\usepackage{rotating}
\usepackage{caption}

\usepackage{algorithm}
\usepackage{algorithmic}
\usepackage{arydshln}
\usepackage{xcolor}
\usepackage{amsthm}
\usepackage[thicklines]{cancel}
\usepackage{ulem}
\makeatletter
\newcommand{\setword}[2]{\textbf{#1}\def\@currentlabel{#1}\label{#2}}
\makeatother

\begin{document}

\title{Scalable Distance-based Multi-Agent Relative State Estimation via Block Multiconvex Optimization  }


\author{\authorblockN{Tianyue Wu, Zaitian Gongye, Qianhao Wang, and Fei Gao}
\authorblockA{Institute of Cyber-Systems and Control,\\ Zhejiang University, Hangzhou 310027 \\ Email: \{tianyueh8erobot, fgaoaa\}@zju.edu.cn}}


%

\maketitle

\begin{abstract}
This paper explores the distance-based relative state estimation problem in large-scale systems, which is hard to solve effectively due to its high-dimensionality and non-convexity. In this paper, we alleviate this inherent hardness to simultaneously achieve scalability and robustness of inference on this problem. Our idea is launched from a universal geometric formulation, called \emph{generalized graph realization}, for the distance-based relative state estimation problem. Based on this formulation, we introduce two collaborative optimization models, one of which is convex and thus globally solvable, and the other enables fast searching on non-convex landscapes to refine the solution offered by the convex one. Importantly, both models enjoy \emph{multiconvex} and \emph{decomposable} structures, allowing efficient and safe solutions using \emph{block coordinate descent} that enjoys scalability and a distributed nature. The proposed algorithms collaborate to demonstrate superior or comparable solution precision to the current centralized convex relaxation-based methods, which are known for their high optimality. Distinctly, the proposed methods demonstrate scalability beyond the reach of previous convex relaxation-based methods. We also demonstrate that the combination of the two proposed algorithms achieves a more robust pipeline than deploying the local search method alone in a continuous-time scenario.  \vspace{-0.2cm}
\end{abstract}

\IEEEpeerreviewmaketitle

\section{Introduction}
\vspace{-0.1cm}
\label{sec: intro}
Relative state estimation is a fundamental problem in multi-agent systems, where individual agents try to estimate the state of other agents relative to themselves. Although there are notable approaches \cite{nguyen2023relative,fishberg2022multi,shalaby2021relative,cossette2022optimal,xu2020decentralized,tian2022kimera} proposed for solving this problem with a limited number of agents, the exploration of such techniques that can scale to tens or hundreds of agents remains largely uncharted. At the same time, control approaches built specifically for large-scale systems have emerged in recent years \cite{bandyopadhyay2017probabilistic,saravanos2023distributed,sravanos2023distributed}, yet under the protection of external localization devices (e.g., motion capture systems) due to the lack of reliable relative state estimation schemes that provide information about inter-agent collision and collaboration. Therefore, we are motivated to study the relative estimation problem in the context of large-scale systems.

Modern distance sensors like ultra-wideband (UWB) offer a lightweight yet effective choice for inter-agent measurements in large-scale systems: they can measure distances over long ranges, spanning tens to hundreds of meters, and accommodate a large number of tags with unique IDs. Despite the unique physical benefits from distance sensors when deployed for state estimation at scale, the corresponding optimization problem is highly non-convex and  high-dimensional. As a result, inference on this problem is sensitive to initial guesses and time-consuming. To overcome the sensitivity to initialization, semidefinite relaxation (SDR) methods are recently applied to find solutions for relevant problems \cite{nguyen2023relative,halsted2022riemannian,liberti2014euclidean}. Unfortunately, the centralized nature and scale-sensitive complexity prevent such methods from being applied to high-dimensional problems. A challenging question, therefore, is how to exploit the robustness of such relaxation models, while allowing it to be solved in a \emph{scalable} way. Moreover, relaxation-based methods are based on some loose convexification of the problems under noisy distance measurements, i.e., the solutions of the relaxation models are deviated from the global optimum of the original problems. A common approach is to further \emph{refine} the solutions using local search methods based on numerical optimization \cite{liberti2014euclidean,halsted2022riemannian}, which are initialized with the solutions of the relaxation models. However, the highly non-convex landscape and high dimension leads to very different estimation accuracies and efficiency obtained by different choices of local search methods \cite{cossette2022optimal,liberti2014euclidean} (see Section \ref{sec:7A}). Therefore, another important problem is to design a local search method that is both accurate and efficient to refine upstream solutions.

\begin{figure*} \centering
	{\includegraphics[width=2\columnwidth]{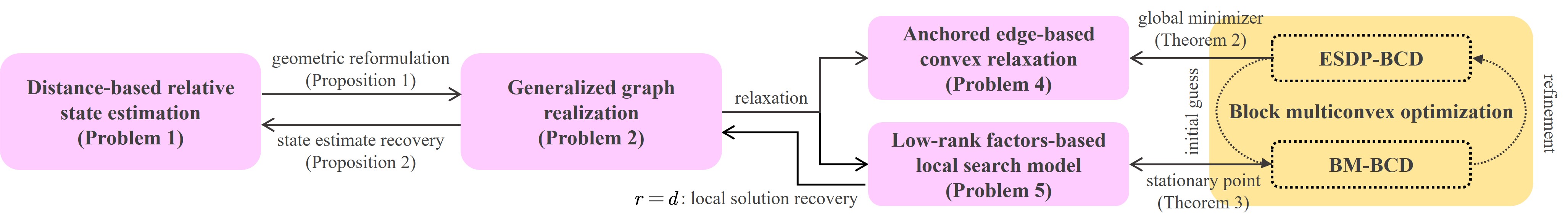} }    
	\vspace{-0.2cm}
	\caption{ \textbf{Relations between problems and algorithms in this work.} The \ref{head} section explains these problems and algorithms and corresponds them to the parts of the article.}
	\label{fig:head}
	\vspace{-0.4cm}      
\end{figure*}

In this paper, we present one suite of such methods to address the above problems in a unified theoretical framework, which contribute two key ideas: (a) a universal formulation, \emph{generalized graph realization}, for distance-proprioception and distance-only relative state estimation; (b) \emph{multiconvex} \cite{xu2013block} and \emph{structurally decomposable} optimization models with solution by \emph{block coordinate descent} (BCD) \cite{bertsekas2015parallel}, which enable both scalable solution of a robust convex relaxation and a following accurate refinement. The key idea (a), generalized graph realization, is the recipe to bridge the distance-based state estimation problem with the idea (b). Its essence lies in a geometric description of rigid body states, inspired by the well known idea that \emph{the state of a rigid body in space can be determined using 3 non-colinear points on it} \cite{horn1987closed}. The key idea (b) includes  a convex relaxation and a local search model, both enjoying decomposable and multiconvex structures, and exploits the scalability of BCD. The decomposable structures allow natural instantiation of the BCD framework. The block multiconvexity \cite{xu2013block} design helps BCD to be instantiated with efficiency and safety, in the sense that each subproblem in the BCD allows fast implementation of exact solutions and convergence of iteration sequence can be formally guaranteed.  In particular, the introduced convex model, unlike the SDR in \cite{nguyen2023relative,liberti2014euclidean,halsted2022riemannian}, is an \emph{edge-based} convex relaxation \cite{waki2006sums,wang2008further} featured by its decomposed constraints. The specialized local search model is used for accurately refining the crude solution by the convex relaxation model. The unique efficiency of this local search method also enables \emph{online estimation} in continuous-time scenarios (see Section \ref{sec:7C}). Thanks to the BCD framework, our approach can be implemented in a distributed manner where each agent performs local computation and communication (see Section \ref{sec:4b}).

Our main contributions are as follows:
\begin{itemize}
	\item [1)] 
	A generalized graph realization formulation for distance-based relative state estimation, supporting both distance-only and distance-proprioception setups.
	\item [2)] 
	Two multiconvex and decomposable optimization models, an edge-based relaxation model for initial guess-robust inference and a local search model for fast refinement; respective instantiations of the BCD framework on the two models, which result in two collaborative algorithms that are scalable and proved to converge in a unified theoretical framework.   
	\item [3)] 
	Benchmark results of the proposed methods on large-scale problems with various sensor and system setups, which show that (i) the proposed two optimization methods collaborate to provide solutions with better or comparable precision against alternative methods, and (ii) the proposed methods exhibit scalability that other convex relaxation methods are naturally unavailable. 
	\item [4)]
	Demonstration of the convex relaxation-based method in a continuous-time scenario, where only high-frequency estimation by the local search method is not enough to provide time-consistent estimates.
\end{itemize}

\setword{Technical Roadmap}{head}\textbf{.} Fig. \ref{fig:head} illustrates the relations between problems and algorithms in this work. Section \ref{sec:3} mainly presents the transformation from the original problem (Problem \ref{problem 1}) to the equivalent generalized graph realization problem (Problem \ref{problem 2} and Proposition \ref{proposition 1}), which is one of our main contributions. Section \ref{sec:4} reviews the BCD algorithm, especially BCD on block multiconvex optimization that is used in this paper. Section \ref{sec:5} derives the optimization models derived from Problem 2 that BCD finally solves, which forms the first half of contribution 2). Section \ref{sec:6} presents the implementation of BCD on the two optimization models, which leads to two collaborative algorithms, BM-BCD and ESDP-BCD, with convergence guarantees of their iteration sequences (Theorems \ref{theorem 2} and \ref{theorem 3}) in a unified theoretical framework (Theorem \ref{theorem 1}). These results form the second half of contribution 2). \vspace{-0.1cm}

\section{Related Work}
\vspace{-0.05cm}
\subsection{Relative State Estimation}
\vspace{-0.1cm}
A lot of previous works have studied the problem of relative state estimation within a few agents. Fusion of intra-agent (e.g., odometry) and inter-agent measurements is a popular paradigm for relative state estimation \cite{nguyen2023relative,xu2020decentralized,tian2022kimera}. \citet{nguyen2023relative} propose QCQP and SDP formulations for the relative transformation estimation problem in 4-DoF (assuming known roll and pitch angles measurements by IMUs) of VIO-UWB fusion.  \citet{ziegler2021distributed} are one of the few examples that employ distributed optimization techniques (in their work, ADMM) for scalably estimating relative states, where the setup is similar to \cite{nguyen2023relative}. However, the good performance of their method is based on the assumption that all agents are equipped with accurate odometry to provide near-perfect initial guesses.  

Some works \cite{shalaby2021relative,cossette2022optimal,fishberg2022multi} propose to equip each agent with multiple distance sensors to achieve an odometry-free relative localization approach. These methods minimize the data throughput requirements and, with an elaborate calibration of the noise distribution, this setup can even achieve similar accuracy to those using odometry \cite{fishberg2022multi}. Although these works consider similar setups to ours, they have yet to explicitly address the computational challenges arising in the context of large-scale systems.  The techniques introduced in this paper can be directly applied to the setups in those works and thus contribute in a complementary way with them. \vspace{-0.2cm}
\subsection{Semidefinite Programming for State Estimation} 
\vspace{-0.1cm}
\subsubsection{Pose Graph Optimization}
Recently, SDP has been successfully applied to the Pose Graph Optimization (PGO) problem. SE-Sync \cite{rosen2019se} and DC2-PGO \cite{tian2021distributed} (a distributed variants of SE-Sync) employ the \emph{Riemannian Staircase} (RS) approach \cite{boumal2015riemannian} that allows to solve SDP efficiently using nonlinear programming (NLP) algorithms. In particular, DC2PGO uses a BCD-type distributed solver \cite{tian2019block} to solve the NLP generated by RS. Unfortunately, RS is not applicable to our problem since the semidefinite matrix in our formulation (see (\ref{eq:Z>0})) does not enjoy the distinctive geometry as those in \cite{boumal2015riemannian}, which is necessary for the adoption of RS. Instead, we employ a structurally decomposable relaxation model where the BCD framework can be applied naturally.
\subsubsection{Distance-based Robotic Localization}
Unlike in PGO problems, where vanilla rank relaxation methods can exhibit \emph{guaranteed} tightness \cite{bandeira2017tightness}, this is often not the case in problems containing noisy distance measurements \cite{so2007theory}.  \citet{halsted2022riemannian} propose \emph{Riemannian Elevator} to identify a rank-restricted space guaranteeing the existence of solutions for Shor relaxation. \citet{papalia2023certifiably} employ an improved certification procedure of RS \cite{rosen2021scalable} to find a solution for the Shor relaxation of distance-aided SLAM. One crucial feature of these approaches is the introduction of edge directions \cite{halsted2022riemannian} that leads to homogeneous QCQPs over nonlinear Riemannian manifolds. In parallel, \citet{dumbgen2022safe} derive the optimality certificate for the anchored, distance-only, and single-agent setup using a quadratic measurement model as in this paper. Interestingly, they empirically exhibit tightness of the relaxation at low noise levels. In a subsequent work \cite{dumbgen2023toward}, the authors develop a methodological approach to automatically find redundant constraints and ensure the tightness for SDR. In this paper, instead of focusing on the general tightness of SDR, we explore a further relaxation of the SDR \cite{wang2008further} and provide some observation of practical significance of its tightness, as shown in Section \ref{sec:5C}.

Another Guass-Seidel-type method similar to BCD, named \emph{sequential greedy algorithm} (SGA), is used in multi-agent applications \cite{corah2017efficient} and similar distance-based localization problems to ours \cite{shi2010distributed}. As the name of the algorithm suggests, SGA is sequential in nature. In this paper,  we use a coloring scheme to parallelize the computation. More importantly, we exploit multiconvex structures embedded in the formulations to establish global convergence of the sequences, while such a crucial property is \emph{not} formally established for SGA. 

To conclude this section, we note that some recent work on inverse kinematics \cite{maric2021riemannian,giamou2022convex} also explore a point-based description and geometric perspective to characterize the problem and employ a point-based reformulation, using centralized local search \cite{maric2021riemannian} or SDR \cite{giamou2022convex} to achieve state-of-the-art performance. \vspace{-0.05cm}

\section{Problem Statement and Reformulation}
\vspace{-0.05cm}
\label{sec:3}
In this section, we specify the setup of distance-based relative state estimation and transform it into the generalized graph realization problem, which can be compactly formulated as a rank-constrained SDP. \vspace{-0.2cm}

\subsection{Problem Setup}
\vspace{-0.1cm}
We consider a practical multi-agent system where each individual is equipped with one or multiple distance sensors and accessible to certain proprioceptive measurements (e.g., 6-axis or 9-axis inertial measurement units (IMUs)\footnote{References to IMUs in the following text are 4-axis IMUs by default.}). We follow a high-level description of proprioception measurements that an individual agent can obtain an accurate estimate of some components of its attitude, i.e., its roll, pitch and yaw angles, which is also used in related work \cite{nguyen2023relative,shalaby2021relative,ziegler2021distributed}. 

We define a \emph{common reference system}, in which the agents’ states are expressed. Since we are considering the problem of relative state estimation, the common reference system can be set arbitrarily when no proprioception is  employed. Its z-axis or x-axis should be aligned with the opposite direction of gravity or geographic east, respectively, if IMUs or magnetometers (typically embedded in 9-axis IMUs) are used. The goal of this paper is to recover the states of agents in the common reference system, which best explain the given distance-only or distance-proprioception measurements. \vspace{-0.2cm}

\subsection{Notions and Preliminaries}
\label{sec:3b}
\vspace{-0.1cm}
We model the measurement topology underlying the relative state estimation problem using an undirected graph $\mathcal{G} =\left( \mathcal{V} ,\mathcal{E} \right)$ in which each node $i\in \mathcal{V}$ represents an agent in the system and the edges\ $(i,j) \in \mathcal{E}$ indicates the measurements between \mbox{agents $i$, $j$}. An index set $\mathcal{B}_i \coloneqq \left\{ 0,1,...   \right\} $ is introduced to distinguish between the distance sensors on the agent $i$. Let \mbox{$\boldsymbol{p}\coloneqq \left[ p_{1}^{0},...p_{1}^{\left| \mathcal{B} _1 \right|},...,p_{n}^{0},...,p_{n}^{\left| \mathcal{B} _n \right|} \right] \in \mathbb{R} ^{d\times \sum_i{\left| \mathcal{B} _i \right|}}$}
denote a realization of points in $\mathbb{R} ^d$, where $d=3$ by default\footnote{The following results can be easily adapted to the 2-dimensional case.}, and $n$ the number of agents. Each component  $p_{i}^{u}\coloneqq \left[ x_{i}^{u},y_{i}^{u},z_{i}^{u} \right] ^T$
 for \mbox{$i=1,2,...,n$} and $u\in \mathcal{B}_i$, is the coordinate of sensor $(i,u)$. Letting $\tilde{\left( \cdot \right) }$ denote measurements, $\underline{\left( \cdot \right) }$ denote the (latent) ground-truth, we have the following measurement model for the distance between sensors $(i,u)$ and $(j,v)$: 
\vspace{-0.1cm}
\begin{equation}
	\begin{split}
		\label{eq:measurement model0}
		\tilde{d}_{ij}^{uv}=\underline{d}_{ij}^{uv}+\epsilon=\big\| \underline{p}_{i}^{u}-\underline{p}_{j}^{v} \big\| +\epsilon,  \hspace{0.2cm}\epsilon \sim \mathcal{N} \left( 0,\sigma^2 \right), \\ 
		\forall \left( i,j \right) \in \mathcal{E},\ u,v\in \mathcal{B}_i,
	\end{split}	 
	\vspace{-0.3cm}
\end{equation} 
where $\epsilon$ is the measurement noise and $\sigma$ is the noise level.

In this paper, we introduce a quadratic surrogate measurement \cite{trawny2010global}  based on the physical measure model (\ref{eq:measurement model0}), which is ultimately used in optimization: \vspace{-0.1cm}
$$
\label{eq:measurement model^2}
\tilde{q}_{ij}^{uv}\coloneqq\left( \tilde{d}_{ij}^{uv} \right) ^2-\sigma ^2\simeq \left( \underline{d}_{ij}^{uv} \right) ^2+\epsilon _{ij}^{uv}, \epsilon _{ij}^{uv}\sim \mathcal{N} \left( 0,(\sigma_{ij}^{uv})^2\right),    
$$
\vspace{-0.4cm}
\begin{equation}
	\sigma_{ij}^{uv}=\sqrt{\big(2\sigma \tilde{d}_{ij}^{uv} \big) ^2+2\sigma ^4}, \ \ 
	\forall \left( i,j \right) \in \mathcal{E},\ u,v\in \mathcal{B}_i.
\end{equation}

Letting $\underline{R}_i\in \mathrm{SO}\left( d \right) $, $\underline{t}_i\in \mathbb{R} ^d$
represent the rotation and translation components of the agent $i$'s state in the common reference system and $\bar{\nu}_{i}^{u}$ the (known) coordinate of \mbox{sensor $u$} on the agent $i$ in its body reference system, we have \vspace{-0.1cm}
\begin{equation}
	\label{eq:p=Rv+t}
	\underline{p}_{i}^{u} = \underline{R}_i\bar{\nu}_{i}^{u}+\underline{t}_i.
\vspace{-0.1cm}
\end{equation}
Given a set of noisy measurements $\tilde{q}_{ij}^{uv}$, it is straightforward to derive the \mbox{maximum-likelihood} estimation (MLE) model for distance-based relative state estimation: \vspace{0.1cm} \\ 
\textbf{Problem \setword{1}{problem 1} \textnormal{(The MLE model for distance-based relative state estimation)}.}
\vspace{-0.2cm}
\begin{equation}
	\label{eq:cost}
	\hspace{-0.4cm}
	\underset{\left\{ (R_i, t_i) \right\}}{\min}\hspace{-0.2cm}\hspace{-0.1cm}\sum_{\footnotesize \begin{array}{c}
			\left( i,j \right) \in \mathcal{E}\\
			u,v\in \mathcal{B}_i\\
	\end{array}}\hspace{-0.2cm}{\hspace{-0.2cm}\frac{1}{ (\sigma _{ij}^{uv})^2 }\left( \left\| R_i\bar{\nu}_{i}^{u}+t_i-R_j\bar{\nu}_{j}^{v}-t_j \right\|^2 \hspace{-0.15cm}-\tilde{q}_{ij}^{uv} \right) ^2}\hspace{-0.1cm},
	\vspace{-0.3cm}
\end{equation}
where \emph{some components of $R_i$ is measured via proprioception}.  \vspace{-0.5cm}

\subsection{From Relative State Estimation to Graph Realization}
\vspace{-0.1cm}
\label{sec:3C}
In this subsection, we transform Problem 1 to a problem that finds the realization of distance sensors best explaining the measurements and rigid body constraints, bridging Problem 1 with the optimization models derived in Section \ref{sec:5}. 

The core idea of this approach lies in the use of a set of points on a rigid body to represent its rotation \cite{horn1987closed}, degrading the manifold constraints imposed by rotations to intuitive \emph{calibration and proprioception constraints}, as follows: \vspace{0.1cm} \\
\textbf{Problem \setword{2}{problem 2} \textnormal{(Generalized graph realization (with reduntant constraints))}.} \vspace{-0.3cm}
\begin{equation}
	\label{eq:problem 2}
	\underset{\boldsymbol{p}}{\min}\sum_{\footnotesize \begin{array}{c}
			\left( i,j \right) \in \mathcal{E}\\
			u,v\in \mathcal{B}_i\\
	\end{array}}{\frac{1}{ (\sigma _{ij}^{uv})^2 }\left( \left\| p_{i}^{u}-p_{j}^{v} \right\|^2 -\tilde{q}_{ij}^{uv} \right) ^2},
	\vspace{-0.4cm} 
\end{equation}
 s.t. \vspace{-0.1cm}
 \begin{itemize}
 	\item Distance-only setup where all the distance sensors are \emph{coplanar}:  
\end{itemize}
 	\begin{equation}
 		\label{eq:norm constraint}
 		\left\| p_{i}^{u}-p_{i}^{v} \right\|^2 =\left\| \bar{\nu}_{i}^{u}-\bar{\nu}_{i}^{v} \right\|^2, \ \forall i\in \mathcal{V} ,\left| \mathcal{B} _i \right|\geq 2, u,v\in \mathcal{B}_i , u< v; 
 		\vspace{-0.1cm}
 	\end{equation}
\begin{itemize}
 	\item distance-4-axis-proprioception setup: 
\end{itemize}
 	\begin{equation}
 		\label{eq:z constriant}
 		\begin{split}		
 			z_{i}^{u}-z_{i}^{v}=\left[ \begin{matrix}
 				-\sin \tilde{\phi}_i,\	\cos \tilde{\phi}_i\sin \tilde{\theta}_i, \		\cos \tilde{\phi}_i\cos \tilde{\theta}_i\\
 			\end{matrix} \right] \left( \bar{\nu}_{i}^{u}-\bar{\nu}_{j}^{v} \right)\hspace{-0.05cm}, 
 		\end{split}
 	\end{equation}
 	\vspace{-1.7cm}
 	\begin{center}
 		$$	\forall i\in \mathcal{V}, \left| \mathcal{B} _i \right|\geq 2, u,v\in \mathcal{B}_i , u< v,$$
 	\end{center}
 	\vspace{-0.8cm}
 	\begin{center}
 		\footnotesize$${\left( \bar{\nu}_{i|x}\left( y_{i}^{u}-y_{i}^{v} \right) -\bar{\nu}_{i|y}\left( x_{i}^{u}-x_{i}^{v} \right) \right)}={\left( \bar{\nu}_{i|x}\left( y_{i}^{v}-y_{i}^{w} \right) -\bar{\nu}_{i|y}\left( x_{i}^{v}-x_{i}^{w} \right) \right)}, $$
 	\end{center}
 	\vspace{-0.8cm}
 	\begin{center}
 		\footnotesize$${\left( \bar{\nu}_{i|x}\left( x_{i}^{u}-x_{i}^{v} \right) +\bar{\nu}_{i|y}\left( y_{i}^{u}-y_{i}^{v} \right) \right)} = {\left( \bar{\nu}_{i|x}\left( x_{i}^{v}-x_{i}^{w} \right) +\bar{\nu}_{i|y}\left( y_{i}^{v}-y_{i}^{w} \right)\right)},$$
 	\end{center}
 	\vspace{-0.12cm}
 	\begin{equation}
 		\label{eq:yaw}
 		\forall i\in \mathcal{V}, \left| \mathcal{B} _i \right|\geq 3, u,v,w\in \mathcal{B}_i , u< v<w, \vspace{-0.1cm}
 	\end{equation}
 where $\left[ \bar{\nu}_{i|x};\bar{\nu}_{i|y};\bar{\nu}_{i|z} \right]\coloneqq \hat{R}_{i|y}\hat{R}_{i|x}\left( \bar{\nu}_{i}^{u}-\bar{\nu}_{i}^{v} \right)$, and  $\hat{R}_{i|x}$, $\hat{R}_{i|y}$ are  the estimates from proprioception sensors of x-axis and y-axis rotation components, respectively, in the agent $i$'s attitude;
\begin{itemize}
\item distance-6-axis-proprioception setup:
 \end{itemize} 
 \begin{equation}
 	\label{eq:6 axis IMU}
 	p_{i}^{u}-p_{i}^{v}=\hat{R}_i\left( \bar{\nu}_{i}^{u}-\bar{\nu}_{i}^{v} \right), \ \forall i\in \mathcal{V}, \left| \mathcal{B} _i \right|\geq 2, u,v\in \mathcal{B}_i , u< v, \vspace{-0.1cm},
 \end{equation}
where $\hat{R}_i$ is the estimates of agent $i$ the full attitude from proprioception sensors. \vspace{0.1cm} \\
\textbf{Proposition \setword{1}{proposition 1} \textnormal{(Equivalence between Problem 1 and Problem 2)}.} Problem 1 and Problem 2 are equivalent in the sense that (a) for any feasible $\{(R_i,t_i)\}$ of Problem 1, $\{p_i^u\}$ that is obtained from \vspace{-0.1cm}
\begin{equation}
	\label{eq:equation}
	p_{i}^{u}=R_i\bar{\nu}_{i}^{u}+t_i
	\vspace{-0.1cm}
\end{equation}
is feasible in Problem 2 for all $i\in \mathcal{V}$ and $u\in \mathcal{B} _i$ and leads to the consistent objective function value with Problem 1; and
 (b) for all feasible $\{p_i^u\}$ of Problem 2, there exists $\{(R_i,t_i)\}$, where $R_i$ agrees with the proprioception of agent $i$, such that (\ref{eq:equation})
holds for all $i\in \mathcal{V}$ and $u\in \mathcal{B} _i$, and all $\{(R_i,t_i)\}$ satisfying (\ref{eq:equation}) lead to an objective function value of Problem 1 consistent with that of Problem 2. \vspace{0.1cm}   

We present a high-level insight conveyed by Proposition 1 while deferring the proof of it to Appendix \ref{appendix I}: Proposition 1 suggests that by representing agents' states using point coordinates and preserving constraints related to multiple distance sensors on the same rigid body, i.e., the \emph{calibration constraints} (\ref{eq:norm constraint}) and \emph{proprioception constraints} (\ref{eq:z constriant}) and (\ref{eq:yaw}), we can recover the agents' states by finding a set of points that agrees with these measurements and constraints, which we call a \emph{realization} of the measurements and constraints. \vspace{0.1cm} 

In the following result, we present a method to recover the state ${\left\{ (R_i, t_i) \right\}}$ from $\boldsymbol{p}$, which is held naturally by the proof of Proposition 1. \vspace{0.1cm} \\
\textbf{Proposition 2 \textnormal{(Recovering states from sensor coordinates)}.} 
\begin{itemize}
	 \item With $\left| \mathcal{B} _i \right|\geq 2$ and 4-axis proprioception, we have 
\end{itemize}
$$
\hspace{-0.1cm}\sin \hat{\psi}_i={{\left( \bar{\nu}_{i|x}\left( \hat{y}_{i}^{u}-\hat{y}_{i}^{v} \right) -\bar{\nu}_{i|y}\left( \hat{x}_{i}^{u}-\hat{x}_{i}^{v} \right) \right)}/{( {\bar{\nu}_{i|x}}^2+{\bar{\nu}_{i|y}}^2)}},
$$
\vspace{-0.4cm}
$$
\hspace{-0.1cm}\cos \hat{\psi}_i={{\left( \bar{\nu}_{i|x}\left( \hat{x}_{i}^{u}-\hat{x}_{i}^{v} \right) +\bar{\nu}_{i|y}\left( \hat{y}_{i}^{u}-\hat{y}_{i}^{v} \right) \right)}/{( {\bar{\nu}_{i|x}}^2+{\bar{\nu}_{i|y}}^2 )}}, \vspace{-0.1cm}
$$
where $\hat{\psi}_i$ is the estimate of yaw angle, thus the full attitude $\hat{R}_i$ is recovered. Then we have  $\hat{t}_i=\hat{p}_{i}^{u}-\hat{R}_i\bar{\nu}_{i}^{u}$ that recovers the translation component; 
\begin{itemize}
	\item with $\left| \mathcal{B} _i \right|\geq 3$, we register $\hat{p_{i}^{u}}$, $\hat{p_{i}^{v}}$, $\hat{p_{i}^{w}}$, where $u<v<w$ and $p_{i}^{u}$, $p_{i}^{v}$, $p_{i}^{w}$ are non-collinear, to $\bar{\nu}_{i}^{u}$, $\bar{\nu}_{i}^{v}$, $\bar{\nu}_{i}^{w}$. In this way, the transformation in the common reference system, i.e., $( \hat{R}_i,\hat{t}_i )$, can be obtained in closed-form \cite[Section 5]{horn1987closed}.
\end{itemize}
	\textbf{Remark 1 \textnormal{(Reduntant constraints of Problem 2)}.} When $\left| \mathcal{B} _i \right|\geq 3$, part of the constraints in (\ref{eq:norm constraint})-(\ref{eq:6 axis IMU}) may be redundant, i.e., removing some of them does not make the feasible set larger.  We show a result that helps us to find an equivalent subset of these constraints in Appendix \ref{appendix II}. We use this result in our evaluation (Section \ref{sec:7}) for better efficiency. \vspace{0.1cm}
	\\
	\textbf{Remark 2} (Handedness ambiguities in distance-only setup). Note that we impose all the distance sensors on a plane in the distance-only setup. This is because the distance constraint in (6) is not sufficient to discriminate handness realizations of the true realization in a noncoplanar case. We can eliminate this ambiguity by imposing constraints such as 
	\begin{equation}
		\label{eq:handness}
		\begin{split}
			\left( \left( p_{i}^{u}-p_{i}^{v} \right) \times \left( p_{i}^{w}-p_{i}^{x} \right) \right) \cdot \left( p_{i}^{y}-p_{i}^{z} \right) = \\ 
			\left( \left( \bar{\nu}_{i}^{u}-\bar{\nu}_{i}^{v} \right) \times \left( \bar{\nu}_{i}^{w}-\bar{\nu}_{i}^{x} \right) \right) \cdot \left( \bar{\nu}_{i}^{y}-\bar{\nu}_{i}^{z} \right),
		\end{split}
	\end{equation}
	where $(\bar{\nu}_{i}^{u}-\bar{\nu}_{i}^{v})$, $(\bar{\nu}_{i}^{w}-\bar{\nu}_{i}^{x}))$, and $(\bar{\nu}_{i}^{y}-\bar{\nu}_{i}^{z})$ are not coplanar. These constraints are cubic, and extending the results in this work to such constraints seems untrivial, which we view as an important future direction (see Section \ref{sec:conclusion}). \vspace{-0.1cm}

\subsection{Problem~2 as Rank-constrained SDP}
\label{sec:3D}
\vspace{-0.1cm}
In this subsection, we present a equivalent rank-constrained SDP formulation of Problem 2 for more concise derivation of results in Section \ref{sec:5}. To reveal this, variable $X\coloneqq \boldsymbol{p}^T\boldsymbol{p}$ is critically introduced, which is equivalent to \vspace{-0.1cm}
\begin{equation}
	\label{eq:X>ptp}
	X\succeq \boldsymbol{p}^T\boldsymbol{p}, \ \mathrm{rank}\left( X \right) \leq d, 
	\vspace{-0.1cm}
\end{equation}
and thus can be rewritten as \cite{boyd1994linear}
\begin{equation}
	\label{eq:Z>0}
	Z\coloneqq \left[ \begin{matrix}
		X&		\boldsymbol{p}^T\\
		\boldsymbol{p}&		I_d\\
	\end{matrix} \right] \succeq 0,\ \mathrm{rank}\left( Z \right) \leq d.
\end{equation}
In this way, we can represent the square of distances and the  constraints by linear mappings on the components of decision variable $Z$. For the convenience of the derivations in Section \ref{sec:5}, we divide these linear mappings into two categories. In the first category, we use $\mathcal{A} ^{q|o}\cdot X$ and $\mathcal{A} ^{q|c}\cdot X$ to substitute the \emph{q}uadratic distance $\left\| p_{i}^{u}-p_{j}^{v} \right\|^2$ in the \emph{o}bjective function (\ref{eq:problem 2})  and the \emph{c}onstraint (\ref{eq:norm constraint}), respectively. In the second category, we use $
\mathcal{A} ^l\cdot \boldsymbol{p}$ to uniformly capture the remaining \emph{l}inear constraints (\ref{eq:z constriant})-(\ref{eq:6 axis IMU}). $\mathcal{A} ^{q|o}$, $\mathcal{A}^{q|c}$ and $
\mathcal{A} ^l$ are the corresponding linear operators. In this way, Problem 2 can be rewritten as follows:\vspace{0.1cm} \\ 
\textbf{Problem 3} (Rank-constrained SDP formulation of Problem 2)\textbf{.}
\vspace{-0.3cm}
\begin{equation}
	\min_Z 
	\left\|  \mathcal{A} ^{q|o}\cdot X-\tilde{\boldsymbol{q}}  \right\|  ^2 \vspace{-0.3cm}
\end{equation}
\ \ \ \ \ \ \ \ \ \ \ \ \ \hspace{0.1cm} s.t. \vspace{-0.5cm} 
\begin{center}
	$$
	Z\coloneqq \left[ \begin{matrix}
		X&		\boldsymbol{p}^T\\
		\boldsymbol{p}&		I_d\\
	\end{matrix} \right] \succeq 0,\ \mathrm{rank}\left( Z \right) \leq d, \vspace{-0.2cm}
$$
\end{center}
\begin{equation}
	\label{eq:norm constraint A}
	\mathcal{A} ^{q|c}\cdot X=\bar{\boldsymbol{b}}^{q|c}, 
	\vspace{-0.25cm}
\end{equation}
\begin{equation}
	\label{eq:z constraint A}
		\mathcal{A}^l\cdot \boldsymbol{p}=\bar{\boldsymbol{b}}^l,
\vspace{-0.25cm} 
\end{equation}
where $\tilde{\boldsymbol{q}}$ is constituted by stacking measurements $\tilde{q}_{ij}^{uv}$, and $\bar{\boldsymbol{b}}^{q|c}$ and $\bar{\boldsymbol{b}}^l$ are known constant vectors form constraints (\ref{eq:norm constraint})-(\ref{eq:6 axis IMU}).

This problem is non-convex due to the rank constraint and intractable to solve directly for a Karush-Kuhn-Tucker point (KKT point) \cite{sun2017rank}, not to mention the global minimizer. \vspace{-0.1cm}

\section{Block Coordinate Descent for Multiconvex Optimization}
\label{sec:4}
\vspace{-0.1cm}
In this section, we review the classic BCD algorithm employed in this paper, and highlight a class of problems  that are convex in each block of variables although the objective function and the feasible set are non-convex in general. \vspace{-0.2cm} 

\subsection{Block Coordinate Descent}
\vspace{-0.1cm}
We first recap the BCD framework, which optimizes the following problem \vspace{-0.1cm}
\begin{equation}
	\label{eq:opt}
	\min_{\boldsymbol{x}\in \mathcal{X}} f\left( \boldsymbol{x} \right),
	\vspace{-0.1cm}
\end{equation}
where $\boldsymbol{x}\coloneqq \left[ z_1,...,z_n \right]^T$ includes all variables $z_i$, and $\mathcal{X}$ is the feasible set and assumed closed. Letting \emph{block division}  $\{\varPhi _q\}_{q=1,...,p}: \boldsymbol{x}=\left( x_1,...,x_p \right) $ divide $\boldsymbol{x}$ into $p$ disjoint blocks, where each block of variables are donated as $x_q$, the BCD algorithm of Gauss-Seidel type \cite{bertsekas2015parallel} optimizes the following \emph{block update subproblem} over each block $\varPhi _q$ \emph{cyclically} \vspace{-0.2cm}
\begin{equation}
	\label{eq:block update problem}
	\min_{x_q\in \mathcal{X} _q}f_q(x_q)\coloneqq 
	 f\left( \hat{x}_1,...,x_q,...,\hat{x}_p \right), \vspace{-0.1cm}
\end{equation}
where $\hat{x}_i$ represents the variables excluded by block $\varPhi _q$ and remains \emph{fixed} at its last updated value in the BCD framework, and \vspace{-0.2cm}
\begin{equation}
	\label{eq:block set}
	\mathcal{X} _q\coloneqq \left\{ x_q:\left( \hat{x}_1,...,x_q,...,\hat{x}_p \right) \in \mathcal{X} \right\}.
	\vspace{-0.1cm}
\end{equation}

In this paper, we also consider the \emph{prox-linear update}  \cite{xu2017globally,zeng2019global} for non-convex optimization, which exhibits better robustness to initialization while taking more time to converge (see Fig.~\ref{fig:NLP}), of which the $k$-th iteration is written as \vspace{-0.2cm} 
\begin{equation}
	\label{eq:prox-linear update}
	\underset{x_q\in \mathcal{X} _q}{\min}\langle \nabla_q f( \hat{x}_{q}^{\left( k-1 \right)} ) ,x_q-\hat{x}_{q}^{\left( k-1 \right)}\rangle +\frac{L_{q}^{\left( k-1 \right)}}{2}\parallel x_q-\hat{x}_{q}^{\left( k-1 \right)}\parallel ^2,
\end{equation}
where $\nabla_q f$ is the block-partial gradient of $f$, $L_{q}^{\left( k-1 \right)}>0$ and $\hat{x}_{q}^{\left( k-1 \right)}$ is obtained by \emph{extrapolation} \vspace{-0.1cm}
\begin{equation}
	\label{eq:momentum update}
	\hat{x}_{q}^{\left( k-1 \right)}=x_{q}^{\left( k-1 \right)}+w_{q}^{\left( k-1 \right)}( x_{q}^{( k-1 )}-x_{q}^{\left( k-2 \right)} ) 
	\vspace{-0.1cm}
\end{equation}
with $0\leq w_{q}^{\left( k \right)}\leq \delta \sqrt{{{L_{q}^{\left( k-1 \right)}}/{L_{q}^{\left( k \right)}}}}$, where $\delta<1$ uniformly over all $q$ and $k$. We consider the extrapolation (\ref{eq:momentum update}) since it significantly accelerates the convergence of BCD in our application.   \vspace{-0.1cm}

\begin{figure} \centering 
	\setcounter{subfigure}{0} 
	\hspace{-0.15cm}
	\subfigure[]{
		\raisebox{-0.4ex}
		{\includegraphics[width=0.3\columnwidth]{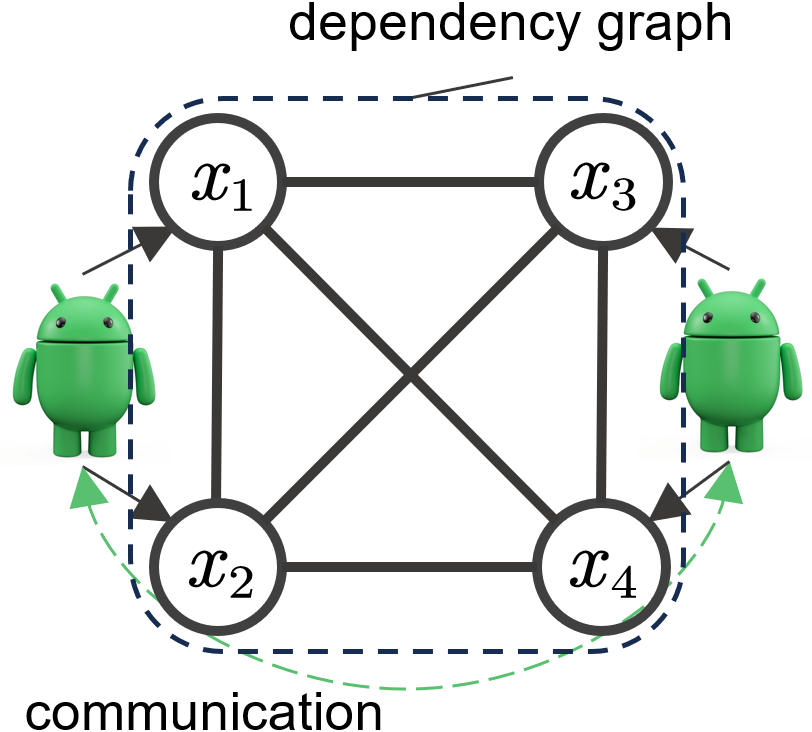}
			\label{fig:dependency graph (a)}}
	}      
	\subfigure[]{  
		\raisebox{0.1ex}  
		{\includegraphics[width=0.3\columnwidth]{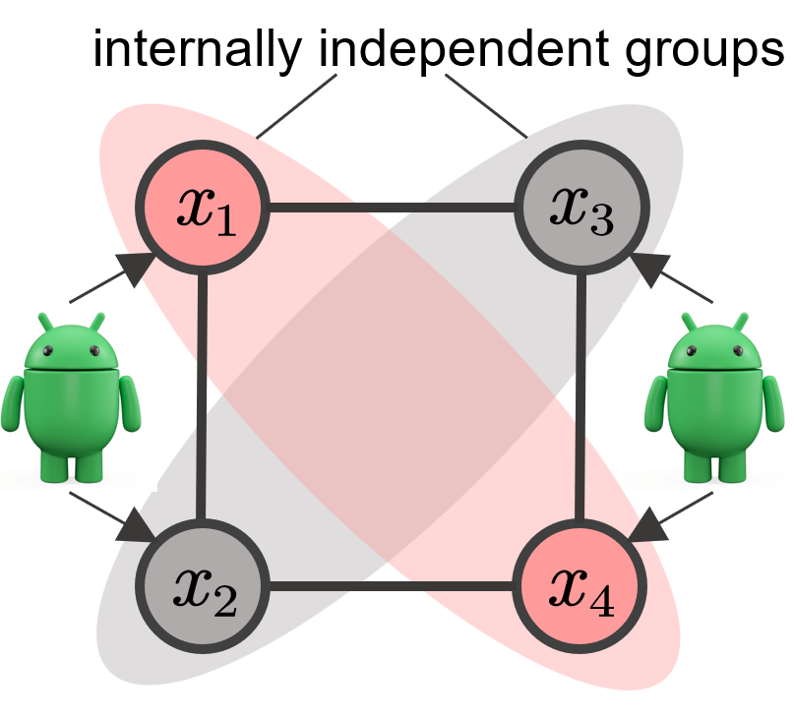}
			\label{fig:dependency graph (b)}}     
	} \vspace{-0.2cm} \\  
	\captionsetup{labelformat=simple}
	\captionsetup{font=footnotesize}  
	\caption{\textbf{An example of dependency graphs and  graph-coloring.} (a) A dependency graph of a non-parallelizable case, where each agent is responsible for two variables, and the solid lines indicate the dependency relations between variables. (b) A dependency graph of a parallelizable case and a feasible coloring scheme of the blocks.}   
	\vspace{-0.6cm}
	\label{fig:dependency graph}
\end{figure}

\subsection{The Scalability and Distributed Nature of BCD}
\label{sec:4b}
\vspace{-0.05cm}
BCD is naturally suited to distributed architectures and is scalable, which we highlight in the following two ways.

\textbf{Parallelization.} The above updates (\ref{eq:block update problem}) and (\ref{eq:prox-linear update}) can be uniformly represented as a general \emph{single-step Gauss-Seidel iteration}, called a \emph{sweep} \cite[Chapter 1]{bertsekas2015parallel}:  
\begin{equation}
	\label{eq:sweep}
	x_{q}^{\left( k+1 \right)}=\pi _q( x_{1}^{\left( k+1 \right)},\cdots ,x_{q-1}^{\left( k+1 \right)},x_{q}^{\left( k \right)},\cdots ,x_{p}^{\left( k \right)} ),
	\vspace{-0.1cm}
\end{equation}
where $\pi_q$ is a mapping function for update w.r.t. block $\varPhi _q$. We introduce the notion of \emph{dependency} as follows: \vspace{0.1cm} \\
\textbf{Definition 1} (Dependency between blocks) Given the arbitrary block division $\{\varPhi _q\}$ and sweep $\pi_q$ for $q=1,...,p$, the block variable $x_j$ is called \emph{independent} on variable $x_k$ if the result of corresponding sweep $\pi_j$ is \emph{invariant} under different values of $x_k$. Conversely, we say that $x_j$ \emph{depends} on $x_k$. \vspace{0.1cm} 

A graphic language describing the dependency relationships between blocks is called \emph{dependency graph} \cite{bertsekas2015parallel,tian2021distributed}, of which we show examples in Fig. \ref{fig:dependency graph}. A sweep (\ref{eq:sweep}) can be non-parallelizable if every block depends on all other blocks (see Fig. \ref{fig:dependency graph (a)}). In contrast, in our problem formalized in Section \ref{sec:3b}, an agent usually generates measurements only with those sensors on a subset of other agents, called the \emph{neighbors} of the agent. We can tell from (\ref{eq:cost}) or (\ref{eq:problem 2}) that a variable $p_i^u$ only depends on partial variables (e.g., see Fig. \ref{fig:dependency graph (b)}) corresponding to sensors on the neighbors of agent $i$. In such a case, parallel computation of sweeps can happen.

A well-known method for identifying \emph{internally independent} groups of blocks for parallelization, where blocks within the same group are independent of each other,  is the \emph{graph-coloring} method \cite{bertsekas2015parallel,tian2021distributed}. Through this method, each variable block in the dependency graph is assigned a color, and the blocks with the \emph{same} color can be merged as a new block to be updated in parallel.

\textbf{Scale-invariant computation and local communication.} The computation demand for each sweep w.r.t. a block is related only to the degree of a block in the dependency graph. Similarly, constructing the sweep problem (\ref{eq:sweep}) only requires the responsible agent for a block to communicate locally with agents corresponding to the neighboring blocks (see Fig. \ref{fig:dependency graph (a)}). Therefore, one can control the degree of the dependency graph to be constant as the system scales up, such that the computation and communication overhead produced by each sweep are scale-invariant. Furthermore, when parallelization is enabled with certain choice of graph-coloring algorithms \cite{barenboim2013distributed}, the number of colors can be easily controlled constant as the system's scale increases. In such a case, the sequential solution complexity per cycle of BCD, which is linearly related to the number of colors, can be bounded, and thus the computation time per cycle in BCD is approximately independent of the system's scale (see Fig. \ref{fig:scalability evaluation}). \vspace{-0.15cm}

\subsection{Convergence of BCD in Multiconvex Optimization}
\vspace{-0.1cm}
Although the BCD algorithm has many desired properties, it does not operate safely in general problems \cite{wright2015coordinate} where its convergence is often not guaranteed. Therefore, we take advantage of a specific structure, the \emph{block multiconvex} structure \cite{xu2013block}, to wisely develop safe BCD-type algorithms.

The term of \emph{block multiconvex} is formalized as follows. \vspace{0.1cm} \\
\textbf{Definition 2 \textnormal{(Block multiconvexity \cite{xu2013block})}.} We call a set $\mathcal{X}$ \emph{block multiconvex} if its projection to each block of variables, i.e.,  $\mathcal{X} _q$ (\ref{eq:block set}), is convex. We call a function $f$ block multiconvex if $f$ is a convex function for each block of variables while other blocks are fixed. \vspace{0.1cm} 

When $\mathcal{X}$ and $f$ are both block multiconvex, the subproblems in updates (\ref{eq:block update problem}) and (\ref{eq:prox-linear update}) are convex. We also need the term \emph{Nash equilibrium} to describe the convergence property. \vspace{0.1cm}\\
\textbf{Definition 3 \textnormal{(Nash equilibrium)}.} We call $\boldsymbol{x}^*=\left( x_{1}^{*},...,x_{p}^{*} \right)$ a \emph{Nash equilibrium} of (\ref{eq:opt}) if \vspace{-0.1cm}
\begin{equation}
	\label{eq:Nash equilibrium}
	f\left( x_{1}^{*},...,x_{q}^{*},...x_{p}^{*} \right) \leq f\left( x_{1}^{*},...,x_q,...x_{p}^{*} \right), \ \forall x_{q} \in \mathcal{X} _q, 
	\vspace{-0.1cm}
\end{equation}
holds for all $q=1,...,p$.  \vspace{0.1cm} 

In the following Theorem 1, we present a crucial result applied in this paper, which claims the (asymptotical) progress of the sequence generated by BCD to a Nash equilibrium. \vspace{0.1cm} \\
\textbf{Assumption 1 \textnormal{\cite[Assumption 2]{xu2013block}}.} $f(\cdot)$ is continuous, bounded, and has a Nash equilibrium. 
\begin{itemize}
	\item The problem is strongly block multiconvex under the block division scheme, i.e., the subproblem is \emph{strongly convex}, when the original update (\ref{eq:block update problem}) is employed;
	\item $\nabla_q f$ is \emph{Lipschitz continuous} and there exist parameters $L_q^{(k-1)}$ obey $\footnotesize f_q( x_{q}^{\left( k \right)} ) -f_q( \hat{x}_{q}^{\left( k-1 \right)} ) \leq \langle \nabla _qf(\hat{x}_{q}^{\left( k-1 \right)}),x_q-\hat{x}_{q}^{\left( k-1 \right)}\rangle +\frac{L_{q}^{\left( k-1 \right)}}{2}\parallel x_q-\hat{x}_{q}^{\left( k-1 \right)}\parallel ^2$, which is used in (\ref{eq:prox-linear update}),
	when the prox-linear update (\ref{eq:prox-linear update}) is employed.
\end{itemize}  
\textbf{Theorem \setword{1}{theorem 1} \textnormal{(Global convergence to a Nash equilibrium \cite[Corollary 2.4]{xu2013block})}.} Under Assumption 1, if  the sequence $\left\{ \boldsymbol{x}^{\left( k \right)} \right\}$ generated by update (\ref{eq:block update problem}) or (\ref{eq:prox-linear update}) is bounded, we have $\lim_{k\rightarrow \infty} \,\mathrm{dist}(\boldsymbol{x}^{\left( k \right)}, \mathfrak{N} 
)=0$, where  $\mathfrak{N}$ is the set of Nash equilibrium.\vspace{0.1cm} \\
We apply this result to establish the convergence guarantee to a \emph{critical point} for the proposed methods in Section \ref{sec:6C}.

\section{Practical Optimization Models for Generalized Graph Realization}
\vspace{-0.1cm}
\label{sec:5}
In this section, we derive a convex model and a local search model based on necessary low-level designs, which are highlighted in \textbf{bold} at the beginning of corresponding paragraphs. We begin by motivating the utilization of ad hoc optimization models instead of Problem 3 and vanilla SDR. \vspace{-0.1cm}

\subsection{Dilemma of Estimating with Problem 3}
\vspace{-0.1cm}
\label{sec:5A}
A classic approach in related literature \cite{so2007theory,liberti2014euclidean,sun2017rank} to general rank-constrained SDP is to drop the rank constraint in (\ref{eq:Z>0}) and obtain a (convex) SDP surrogate, solve the SDP, and rounding a solution of the original problem. However, the constraint (\ref{eq:Z>0}) on $Z$ leads to a polynomial increase in computational complexity with the systems' scale \cite{liberti2014euclidean,halsted2022riemannian}, which cannot be operated scalably with general-purpose solvers \cite{mehrotra1992implementation}. 

Moreover, minimizing the objective function in Problem 3 does not impose any direct optimization on the first-order variable component $\boldsymbol{p}$ of $Z$, and thus $\boldsymbol{p}$ eventually converges to a \emph{trivial} feasible solution. Thus, it is impossible to rounding an informed estimate directly from the (1,2) (first row, second column) or (2,1) (second row, first column) block of $Z$.

An alternative rounding method is to perform approximate rank-$d$ decomposition, e.g., through eigenvalue decomposition, on $X$, the variable component of $Z$ encoded into all the information regarding minimizing the objective function. This approach, although preserving the optimization behavior, cannot maintain the constraints imposed by constraints~(\ref{eq:norm constraint A}) and (\ref{eq:z constraint A}) on the (1,2) or (2,1) block in $Z$.

The solution we present in the following subsection  (i) using a nonlinear programming (NLP) model formulation where rounding is straightforward and (ii) is manually arranging a set of anchors, i.e., sensors whose locations are known as a prior. We observe that, fortunately, a \emph{small number} of anchors is sufficient to extract informed estimates from the (1,2) or (2,1) block of $Z$ for the downstream refinement. \vspace{-0.1cm}

\subsection{An Edge-based Semidefinite Relaxation Model}
\label{sec:5C}
The decision variable of the vanilla SDR is densely constrained by the semidefiniteness of the whole cone matrix $Z$, thus not decomposable. Therefore, we are motivated to further relax the full semidefiniteness constraint to decompose $Z$ according to the measurement topology $\mathcal{E}$.

\textbf{Edge-based relaxation of SDP.} We follow the idea of edge-based decomposition appearing in \cite{wang2008further} to \emph{relax} $Z\succeq 0$ such that the cone matrix $Z$ is decomposed into its principal submatrices, as follows:
\begin{equation}
	\label{eq:ESDR}
	\begin{split}
		\left[ \begin{matrix}
			X_{ii}^{uu}&		X_{ij}^{uv}&		\left( p_{i}^{u} \right) ^T\\
			X_{ji}^{vu}&		X_{jj}^{vv}&		\left( p_{j}^{v} \right) ^T\\
			p_{i}^{u}&		p_{j}^{v}&		I_d\\
		\end{matrix} \right] \succeq 0, \\ \forall \left( i,j \right) \in \mathcal{E} \lor \left( i=j\land \left( u,v \right) \in \mathcal{E} _i \right),
	\end{split}
\end{equation}  
where $X_{ij}^{uv}$ is an abbreviation for the auxiliary variable corresponding to $\left( p_{i}^{u} \right) ^Tp_{j}^{v}$ in (\ref{eq:X>ptp}), and
 $\mathcal{E} _i$ is defined as the set of pairs of sensors on agent $i$ that are taken into account in the calibration constraint (\ref{eq:norm constraint}).

The constraints in (\ref{eq:ESDR}) only appear between pairs of sensors where measurement edges exist in $\mathcal{G}$ (recall Section \ref{sec:3b}) or multiple sensors on the same agent. Hence, the resulting problem is called \emph{edge-based SDP} (ESDP). In this way, the high-dimensional constraint $Z\succeq 0$ (\ref{eq:Z>0}) is decomposed into several low-dimensional constraints. We observe in evaluation (Section \ref{sec:7}) that although ESDP is a further relaxation of SDP, solving it results in a similar accuracy with that of SDP. 

\textbf{Manual anchors.} As mentioned in Section \ref{sec:5A}, directly solving the rank-relaxation problem of Problem 3 leads to a dilemma in extracting estimates. Fortunately, Theorem 2 in \cite{so2007theory}, which is originally proposed for the distance-only setup, can be directly extended to the distance-proprioception measurements, which tells: \emph{in the noiseless case, if there are anchors configured to uniquely localize all distance sensors, $\boldsymbol{p}$ can be estimated \emph{exactly} by solving the rank relaxation of Problem 3 (with modification in formulation to include anchors information) and extracted from the (1,2) block of the solution $Z$.} i.e., the relaxation is tight. Conversely, if sensors cannot be uniquely localized, the relaxation is always loose. 

The above result implies that the information provided by anchors is crucial. Inspired by this, we make the following key observation: even with only a \emph{small number} of imperfectly localized anchors, it is possible to recover informed estimates of $\boldsymbol{p}$ from the (1,2) block of $Z$ via ESDP \emph{if these anchor sensors generate measurements with a sufficient proportion of other sensors.} The following remark clarifies that anchoring some agents in practice is often possible. \vspace{0.1cm} \\ 
\textbf{Remark 2 \textnormal{(Feasibility of manual anchors)}.} In practice, we can anchor a single agent with multiple distance sensors, considering that in the context of \emph{relative} state estimation, one agent can be regarded anchored without any loss of generality, to provide anchor information using calibrated coordinates of the sensors on it (as done in Section \ref{sec:7C}). We can also use existing relative state estimation techniques (e.g., \cite{nguyen2023relative}) to provide (imperfect) relative positions of a small number of agents that serve as anchors for large-scale systems.   \vspace{0.1cm} 

We formalize this anchored ESDP as follows:  \vspace{0.1cm} \\
\textbf{Problem \setword{4}{problem 4}} (Anchored edged-based relaxation for Problem 3)\textbf{.} \vspace{-0.2cm}
\begin{equation}
		\min_Z \big\|  \mathcal{A} ^{q|o}\cdot X-\tilde{\boldsymbol{q}}  \big\|  ^2+ \left\|  \mathcal{A}^a\cdot Z-\tilde{\boldsymbol{q}}^a  \right\|  ^2, \vspace{-0.2cm}
\end{equation}
\ \ \ \ \ \ \ \ \ \ \ \ \ \ \ \ \  \ \ \  \ \ \  s.t. \vspace{-0.2cm}
\begin{center}
	 (\ref{eq:norm constraint A}), (\ref{eq:z constraint A}) and (\ref{eq:ESDR}), \vspace{-0.1cm}
\end{center}
where $\mathcal{A}^a \cdot Z$ is introduced to substitute the term $\left\| p_{i}^{u}-a_m \right\| ^2$, the quadratic distance from any anchor $a_m$, and $\tilde{\boldsymbol{q}}^a$ represents the stacked anchor measurements, similar to Section \ref{sec:3D}.

\textbf{How many anchors and anchor measurements are sufficient?} We define 'sufficient' in the sense that we can extract the exact or (at least) non-trivial estimates of the sensors from the (1,2) block of $Z$ by solving Problem 4. Unfortunately, we cannot yet provide the exact conditions that make the above hold.  Nevertheless, the following result implies a necessary number of anchors. \vspace{0.1cm} \\
\textbf{Proposition 3.} Suppose the measurements are noise-free. If a sensor $(i,u)$ is not uniquely localizable, then there must exist $X_{ii}^{uu}$ and $p_i^u$  in the solution set such that $X_{ii}^{uu}>(p_i^u)^Tp_i^u$.    \vspace{0.1cm} 

The above result, proved in Appendix \ref{appendix III}, suggests that if sensors' positions cannot be uniquely determined, then the relaxation is always loose, and an informed estimate can not be safely obtained by solving Problem 4. For example, as is well known, in a distance-only setup, if the number of anchor sensors is less than $d+1$, then we cannot get a uniquely determined sensor position at most times due to reflection invariance. Although including proprioception information allows the number of anchors required to be further reduced, we still consider it safe to use at least $d+1$ anchor sensors. 

\begin{figure} \centering 
	\setcounter{subfigure}{0} 
	\hspace{-0.3cm}
	\subfigure[]{
		{\includegraphics[width=0.456\columnwidth]{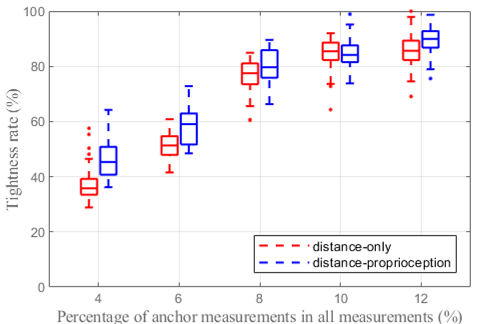}
			\label{fig:tightness rate}}
	}   
	\hspace{-0.3cm}   
	\subfigure[]{  
		{\includegraphics[width=0.452\columnwidth]{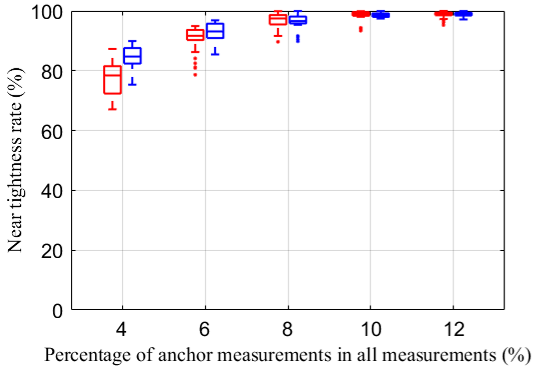}
			\label{fig:near tightness rate}}     
	} \vspace{-0.2cm} \\  
	\captionsetup{labelformat=simple}
	\captionsetup{font=footnotesize}  
	\caption{\textbf{Tightness under varying anchor measurements rate.} The experiments are conducted in a constant topology 3D problem with 125 agents distributed in $[0,12\mathrm{m}]^3$. The agents are either equipped with two distance sensors (the distance-only setup) or also with an additional IMU (the distance-proprioception setup). Two neighboring agents (i.e., four sensors) are used as anchors. (a) The tightness rate which is defined as the proportion of sensors whose localization error is under $10^{-3}$m. (b) The near tightness rate which is defined as the proportion of sensors whose localization error is under $5\times10^{-2}$m.}   
	\vspace{-0.6cm}
	\label{fig:tightness}
\end{figure}

We experiment to determine an empirical amount of anchor measurements required in the case of sparingly using only $d+1$ anchor sensors, of which the results are shown in Fig. \ref{fig:tightness}. Although it is difficult to guarantee absolute tightness with this very limited number of anchors (see Fig. \ref{fig:tightness rate}), we can obtain near-perfect estimates (see Fig. \ref{fig:near tightness rate}) for most agents with a considerably cheap configuration of anchors and anchor measurements. These results validate that a small number of anchor agents can be used to support large-scale ESDP. \vspace{-0.1cm}

\subsection{A Low-Rank Factors-based Local Search Model}
\vspace{-0.05cm}
\label{sec:5B}
In this subsection, we derive a non-convex model with bi-convexity for local search to refine the solution by convex relaxation or perform online estimation. We show in Section \ref{sec:6} that this specialized model leads to favorable computational and convergence properties.

\textbf{Bi-convex factorization.} We draw inspiration from a classic application, \emph{collaborative prediction} \cite{rennie2005fast,srebro2004maximum}, to introduce \emph{bi-convexity} (the 2-blocks instance of block multiconvexity) to the \emph{BM factorization} \cite{burer2003nonlinear} for semidefinite constraint $Z$. By introducing the bi-convexity, we can establish a multiconvex optimization model (as suggested by Lemma \ref{lemma 1}).
 
Collaborative prediction \cite{rennie2005fast,srebro2004maximum} fits an $m\times n$ target matrix $X$ with a pair of rank-$r$ factors $U\in \mathbb{R} ^{m\times r}$ and $V\in \mathbb{R} ^{n\times r}$, such that $X=UV^T$. The key idea to solve it is that if one of the factors, say $U$, is fixed, then fitting the target reduces as  \vspace{-0.2cm}
\begin{equation}
	\label{eq: linear prediction}
	\min_V \big\| \mathcal{A} \cdot \hat{U}V^T-\bar{b} \big\|^{2},
	\vspace{-0.2cm}
\end{equation}
which is \emph{quadratic} on the block variable $V$ and thus can be solved in a closed form. A 2-blocks BCD iterating between $U$ and $V$ can solve the problem.

On the other hand, the proposal of Burer and Monteiro \cite{burer2003nonlinear}, called BM factorization, factorizes a semidefinite positive matrix, say $Z$ in (\ref{eq:Z>0}), with some low-rank matrices $Y\in \mathbb{R} ^{ r\times \sum_i{\left| \mathcal{B} _i \right|}+d}$: \vspace{-0.1cm}
\begin{equation}
	\label{eq:B-M factorization}
	Z=Y^TY=\left[ \begin{array}{c}
		U^T\\
		Q^T\\
	\end{array} \right] \left[ \begin{matrix}
		U& Q\\
	\end{matrix} \right] ,\  Q^TQ=I_d,
	\vspace{-0.1cm}
\end{equation}
where $U\in \mathbb{R} ^{r\times \sum_i{\left| \mathcal{B} _i \right|}}$ and $Q\in \mathbb{R} ^{r\times d}$ with $r$ satisfying \mbox{$d\leq r\leq \sum_i{\left| \mathcal{B} _i \right|}+d$}, being the rank upper bound one imposes on $Z$. This drastically reduces the search space and preserves the semidefiniteness constraint.

Following the above factorizations, we can introduce  bi-convexity into the matrix $Z$ in (\ref{eq:Z>0}) by factorizing it as \vspace{-0.15cm}
\begin{equation}
	\label{eq:BM factorization}
	Z=\left[ \begin{array}{c}
		U^T\\
		Q^T\\
	\end{array} \right] \left[ \begin{matrix}
		V&		Q\\
	\end{matrix} \right] ,\ U-V=O, \ Q^TQ=I_d,
	\vspace{-0.15cm}
\end{equation}  
while preserving the semidefiniteness of $Z$. 

\textbf{Constant projection operator.} The constitutive relation from $U$ ($V$) and $Q$ to $Z$ in (\ref{eq:Z>0}) reveals $Q^TU \equiv \boldsymbol{p}$ ($Q^TV \equiv \boldsymbol{p}$). Consequently, we can view $U$ as an  $r$-dimension surrogate of $\boldsymbol{p}$, while $Q$ acts as a (row) projection operator that maps $U$ onto the realization space $\mathbb{R} ^d$ of $\boldsymbol{p}$.  

Based on the above explanation regarding the physical interpretation of $Q$, we have two options: (i) simultaneously optimizing the projection operator $Q$ and the $r$-dimension realization $U$, and (ii) fixing $Q$ and optimizing $U$ only, as fixing $Q$ does \emph{not} affect on the optimal values of the model and the corresponding optimal estimates. Since fixing $Q$ is expected to result in a more efficient optimization procedure  without the need to preserve the orthogonality constraint, we choose a specific $Q$ and fix it throughout the optimization. Note that with $Q$ fixed, different choices of $Q$ do not intrinsically affect the model's behavior, so we can trivially determine $Q$ as $\left[ \begin{matrix}
	I_d&		{O}\\
\end{matrix}_{d\times \left( r-d \right)} \right]^T$, for example.

\textbf{Surrogate $\boldsymbol{U}$-$\boldsymbol{V}$-coupled constraints.} We replace the constraint (\ref{eq:norm constraint A}) and $U-V=O$ with surrogate functions on the objective function by employing methods such as penalty methods and the augmented Lagrangian method \cite[Chapter 17]{nocedal1999numerical}. These constraints are featured by the fact that $U$ and $V$ are \emph{coupled} in them. The account of necessity for surrogating $U$-$V$ coupled constraints is postponed to Section \ref{sec:6C}. In our implementation, we employ the vanilla quadratic penalty method since it works considerably well in practice.

In comparison, we can rewrite (\ref{eq:z constraint A}) as $\mathcal{A} ^l\cdot \left( Q^TU \right)=\bar{b}^l$ and $\mathcal{A} ^l\cdot \left( Q^TV \right)=\bar{b}^l$ according to $\boldsymbol{p} \equiv Q^TU$ and $\boldsymbol{p} \equiv Q^TV  $, respectively, i.e., $U$ and $V$ are \emph{separate} with each other in constraint (\ref{eq:z constraint A}), and we strictly preserve it as in Problem 3.  

We obtain the following constrained non-convex model:
\vspace{0.1cm} \\
\textbf{Problem \setword{5}{problem 5}} (Low-rank factors-based model for Problem~3)\textbf{.} \vspace{-0.2cm}
\begin{equation}
	\label{eq:cost 2}
	\begin{aligned}
		\min_{U,V} \big\|  \mathcal{A}^{q|o}\cdot (U^TV) -\tilde{\boldsymbol{q}}  \big\|  ^2&+ \\ \lambda \big\|  \mathcal{A} ^{q|c}\cdot (U^TV) -\bar{\boldsymbol{b}}^{q|c}  \big\| ^2 +\gamma &\big\| U-V \big\| _{F}^{2}
		\vspace{-0.3cm}
	\end{aligned}
\end{equation}
\ \ \ \ \ \ \ \ \  \hspace{0.1cm} s.t. \vspace{-0.1cm}
\begin{equation}
	\mathcal{A}^{l}\cdot \left( Q^TU \right) =\bar{\boldsymbol{b}}^l, \vspace{-0.1cm}
\end{equation}
\begin{equation}
	\mathcal{A}^{l}\cdot \left( Q^TV \right) =\bar{\boldsymbol{b}}^l, \vspace{-0.15cm}
\end{equation}
where $\lambda$ and $\gamma$ are the penalty coefficients. 
\vspace{-0.1cm}

\begin{algorithm} 
	\caption{\emph{The Meta Algorithm of Block Coordinate Descent for Problem 4 and Problem 5}} 
	\label{alg} 
	\begin{algorithmic}[1]
		\renewcommand{\algorithmicrequire}{\textbf{Input:}}
		\renewcommand{\algorithmicensure}{\textbf{Output:}}
		\REQUIRE Block division $\{\varPhi _q\}_{q=1,...,p}$, an initialization  $\boldsymbol{p}^{\left( 0 \right)}$       
		\ENSURE Estimates of agents' states $(  \hat{\boldsymbol{R}},\hat{\boldsymbol{t}} ) $
		\STATE $k \gets 0$
		\WHILE {1} 
		\STATE $k \gets k+1$
		\FOR {$q=1,...,p$}
		\STATE Solve (\ref{eq:block update problem}) or (\ref{eq:prox-linear update}) w.r.t. $\varPhi _q$  
		\ENDFOR
		\IF{${\left\| \boldsymbol{p}^{\left( k   \right)}-\boldsymbol{p}^{\left( k-1   \right)} \right\| _F}/{\left\| \boldsymbol{p}^{\left( k-1   \right)} \right\| _F} \leq \epsilon $} 
		\STATE \textbf{break}
		\ENDIF
		\ENDWHILE
		\STATE Recover $( \hat{\boldsymbol{R}},\hat{\boldsymbol{t}} ) $ according to Proposition 2
	\end{algorithmic}
\end{algorithm}
\vspace{-0.3cm}

\begin{figure} \centering 
	\setcounter{subfigure}{0} 
	\subfigure[]{
		\raisebox{-0.8ex}
		{\includegraphics[width=0.28\columnwidth]{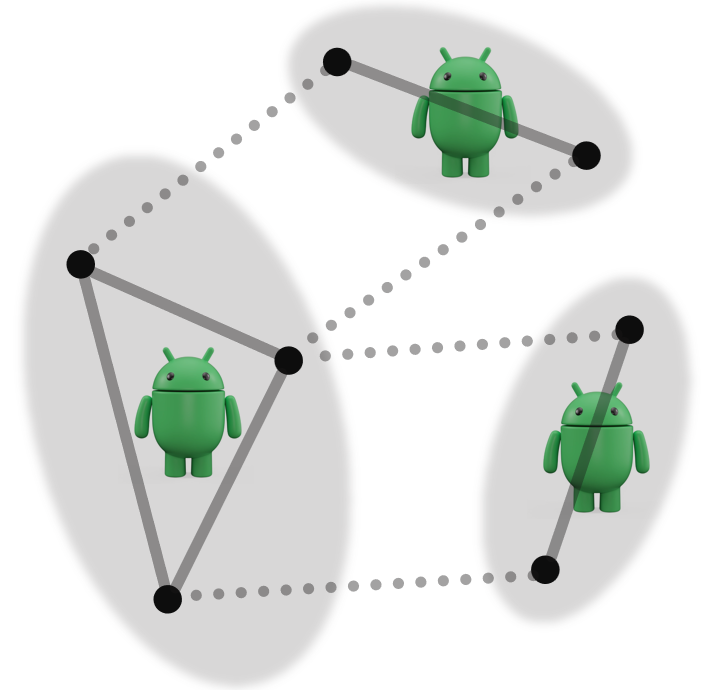}}
	}      
	\hspace{-0.3cm}
	\subfigure[]{   
		\raisebox{0.5ex}
		{\includegraphics[width=0.29\columnwidth]{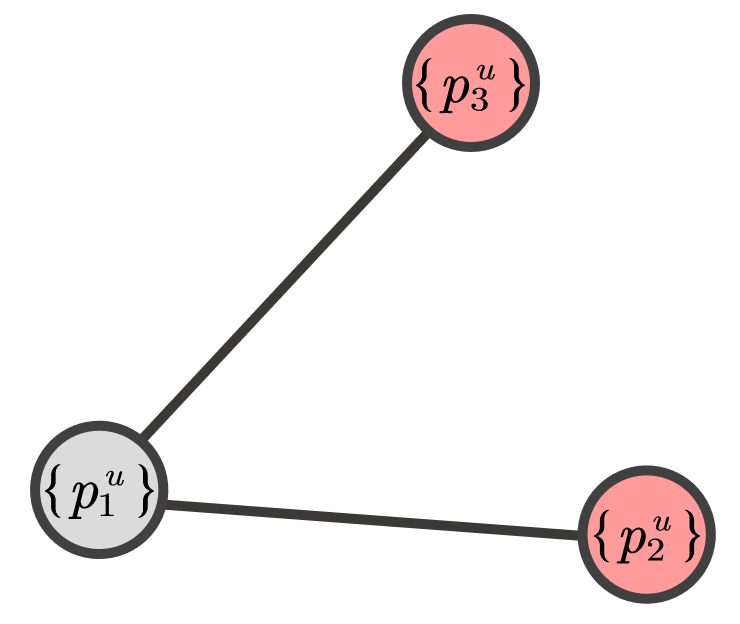}
			\label{fig:block division (c)}}     
	}
	\hspace{-0.4cm}
	\subfigure[]{ 
		\raisebox{0.4ex}  
		{\includegraphics[width=0.33\columnwidth]{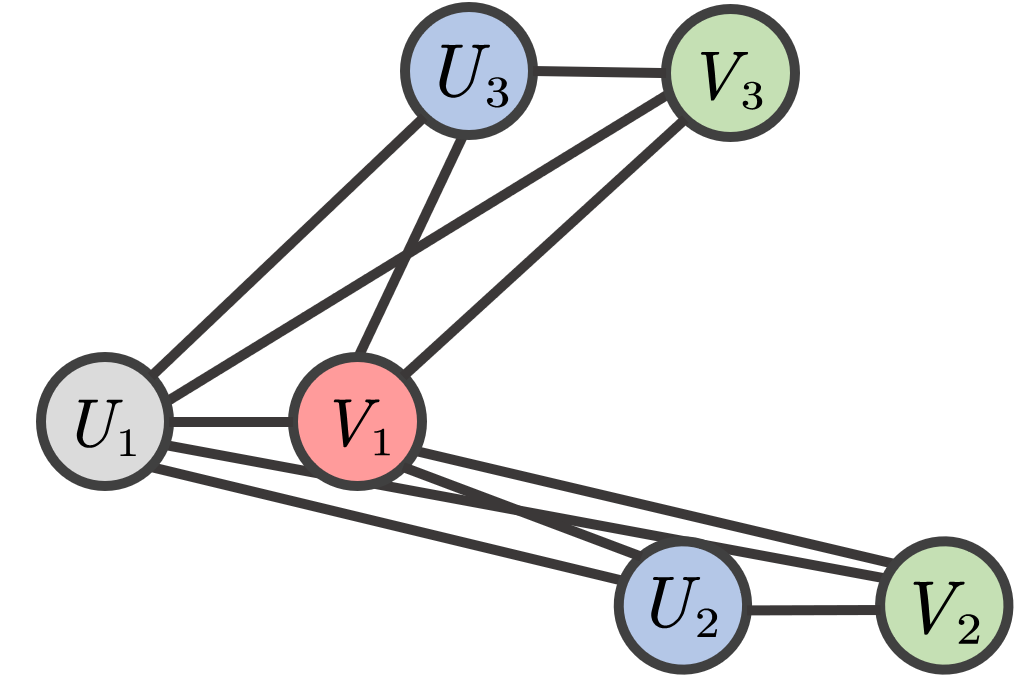}
			\label{fig:block division (b)}}     
	}
	\\
	\vspace{-0.2cm}  
	\captionsetup{labelformat=simple}
	\captionsetup{font=footnotesize}  
	\caption{\textbf{Examples of the specialized block division for Problem 4 and Problem 5, respectively.} (a) An example of a 3-agents system, where the gray ellipses, dashed lines, and solid lines denote the rigid bodies corresponding to agents, distance measures, and rigid body constraints, respectively. (b) The block division for ESDP-BCD in the example and a feasible coloring scheme for it. (c) The block division and a corresponding feasible coloring scheme for BM-BCD in the example, where $U_i$ ($V_i$) represents variables corresponding to the sensors on agent $i$ in $U$ ($V$). }   
	\vspace{-0.6cm}
	\label{fig:block division}
\end{figure}

\section{Solutions via Block Coordinate Descent}
\label{sec:6}
\vspace{-0.05cm}
In this section, we propose solution techniques for Problems 4 and 5, which results in two collaborative algorithms, ESDP-BCD and BM-BCD, with Algorithm \ref{alg} as a unified backbone. In Algorithm \ref{alg}, we need to specify block division for BCD. We establish rules that should be followed by the block division used in the proposed approach, which are necessary for constructing the multiconvex structure that leads to desired computational (Proposition \ref{proposition 4}) and convergence (Theorems \ref{theorem 2} and \ref{theorem 3}) results. \vspace{0.1cm} \\
\textbf{Rule \setword{1}{rule 1}.} No equality constraints exist between variables in different blocks. \vspace{0.1cm} \\
\textbf{Rule \setword{2}{rule 2}.} In Problem 5, variables in $U$ and $V$ cannot be in the same block at the same time.   \vspace{0.1cm} \\
\textbf{Lemma \setword{1}{lemma 1}.} Under Rule 1 and Rule 2, the feasible sets and objective functions of Problem 4 and Problem 5 are block multiconvex and Assumption 1 holds.      \vspace{0.1cm} 

Lemma 1 responds to the requirements of Theorem 1. We apply a simple cyclical block division in the evaluation (Section \ref{sec:7}), as illustrated by Fig. \ref{fig:block division}\textcolor{blue}{:} for Problem 4, all variables corresponding to distance sensors on the same agent constitute a block; for Problem 5, variables in $U$ and $V$ corresponding to distance sensors on the same agent, donated as $U_i$ and $V_i$ in Fig. \ref{fig:block division (b)}, constitute two separate blocks, respectively, which satisfies Rule 2. {Since in Problem \ref{problem 2}, the equality constraints only come from pairs of sensors on the same agent, Rule 1 holds for the above block division schemes.}  Then, we can apply graph-coloring to enable parallel execution, as illustrated in Fig. (\ref{fig:block division (c)}) and Fig. (\ref{fig:block division (b)}).    \vspace{-0.1cm}

\subsection{ESDP-BCD for Convex Block Updating on Problem 4}
We propose \emph{Edge-based SDP with Block Coordinate Descent} (ESDP-BCD) to globally solve Problem \ref{problem 4}. We begin by presenting the block update subproblem of Problem \ref{problem 4} that allows exact and efficient block update and global convergence to a global minimizer (Theorem \ref{theorem 2}).

\textbf{Subproblems of ESDP.} We present the formulation of the subproblem using update (\ref{eq:block update problem}) corresponding to agent $i$ in the following, and postpone its derivation to Appendix \ref{appendix III}. \vspace{-0.5cm} 

$$
\min_{\{p_i^u\},\{X_{ii}^{uu}\}} \big\|  \mathcal{A} ^{q|o}\cdot X-\tilde{\boldsymbol{q}}  \big\|  ^2+ \left\|  \mathcal{A}^a\cdot Z-\tilde{\boldsymbol{q}}_a  \right\|  ^2,  \vspace{-0.2cm}
$$
\ \ \ \ \hspace{0.1cm} s.t. \vspace{-0.1cm}
\begin{equation}
	\label{eq:X>p2}
	X_{ii}^{uu} \geq \left\| p_{i}^{u} \right\| ^2, 
	\forall  u\in \mathcal{B}_i,
	\vspace{-0.7cm}
\end{equation}
\begin{center}
	$$
	X_{ii}^{uu}\geq \left\| p_{i}^{u} \right\| ^2+( X_{ij}^{uv}-( p_{i}^{u} ) ^T\hat{p}_{j}^{v} ) ^2/( \hat{X}_{jj}^{vv}-\left\|\hat{p}_{j}^{v}\right\|^2),
	\vspace{-0.2cm}
	$$
\end{center}
\begin{equation}
	\label{eq:edge constraint}
	\forall j\in \mathcal{N}_i, u\in \mathcal{B}_i ,v\in \mathcal{B}_j, \vspace{-0.1cm}
\end{equation}
\begin{equation}
	\label{eq:inner sdp}
	\left[ \begin{matrix}
		X_{ii}^{uu}&		X_{ii}^{uv}&		\left( p_{i}^{u} \right) ^T\\
		X_{ii}^{vu}&		X_{ii}^{vv}&		\left( p_{i}^{v} \right) ^T\\
		p_{i}^{u}&		p_{i}^{v}&		I_d\\
	\end{matrix} \right] \succeq 0,\forall \left( u,v \right) \in \mathcal{E} _i,
\end{equation}
\begin{center}
	constraints corresponding to agent $i$ in (\ref{eq:norm constraint A}), (\ref{eq:z constraint A}),
\end{center}
where $\hat{p}_{j}^{v}$ and $\hat{X}_{jj}^{vv}$ are the fixed variables corresponding to the sensor $(j,v)$.

The above problem is a (convex) quadratic conic programming that can be solved efficiently using off-the-shelf solvers \cite{mehrotra1992implementation}.  The complexity of solving the problem only depends on the agent $i$'s number of neighbors. Similar results can be established on update (\ref{eq:prox-linear update}), except we have no reason to employ it on such a convex problem. 

The following result, proved in Section \ref{sec:6C}, shows that lines 2-10 of Algorithm \ref{alg} converge a global minimizer of Problem \ref{problem 4}. \vspace{0.1cm} \\
\textbf{Assumption 2.} Every agent is connected in $\mathcal{G}$, directly or indirectly, to at least one agent with a sensor $p_i^u$ such that the sequence $\{ \left( p_{i}^{u} \right) ^{\left( k \right)} \} $ is bounded.   \vspace{0.1cm} \\
\textbf{Assumption 3.} The constraints (\ref{eq:edge constraint}) are replaced with surrogate functions (e.g., the barrier function) in the objective function when solving the subproblems.  \vspace{0.1cm} \\ 
\textbf{Theorem \setword{2}{theorem 2}.} Let $Z^{(k)}$ be the $k$-th iteration generated by lines 2-10 in Algorithm \ref{alg} for solving Problem \ref{problem 4}, and $\mathfrak{M}$ the set of global minimizers of Problem \ref{problem 4}. Under Rules 1 to 2 and Assumptions 2 to 3, we have $\lim_{k\rightarrow \infty} \,\,\mathrm{dist}\left( Z^{\left( k \right)}, \mathfrak{M} \right) =0$. \vspace{0.1cm}

Assumption 3 is proposed for a similar reason to surrogating $U$-$V$-coupled constraints in Section \ref{sec:5B}, which is to construct the \emph{Cartesian product} structure under Rules 1 and 2, as explained in detail in the next subsection. Our implementation uses the interior point method that employs the barrier method \cite{mehrotra1992implementation} to fulfill this assumption.

\textbf{Refinement and extracting estimates.} As claimed in Section \ref{sec:5C}, we can directly extract an informed realization from the (1,2) block of $Z$. Despite not relying on initial guesses, ESDP provides only coarse estimates (especially for the rotation components)  under noisy measurements. To further refine the estimates, we employ BM-BCD introduced in the following subsection. \vspace{-0.1cm}

\subsection{BM-BCD for Fast Local Searching on Problem 5}
\label{sec:6B}
We propose \emph{Burer-Monteiro factorization with Block Coordinate Descent} (BM-BCD) to efficiently solve Problem 5. BM-BCD enables closed-form block update (Proposition \ref{proposition 4}) and global convergence (Theorem \ref{theorem 3}).

\textbf{Linear subproblems.} The formulation of Problem \ref{problem 5} allows extremely fast iterations due to the linearity of its subproblems under Rule \ref{rule 2}, which is formalized as follows:  \vspace{0.15cm} \\
\textbf{Proposition \setword{4}{proposition 4} \textnormal{(Linear subproblems and closed-form updates)}.} Under Rule 2, the updates (\ref{eq:block update problem}) and (\ref{eq:prox-linear update}) for Problem \ref{problem 5} are both the (feasible) \emph{linearly constrained linear least square} problems, which allow closed-form solutions. \vspace{0.1cm}

Under the applied block division described previously, the closed-form update only requires solving a low-dimensional linear system, which can be performed extremely fast. 

The following result, proved in Section \ref{sec:6C}, shows that Algorithm \ref{alg} converge to KKT points of Problem \ref{problem 5}. \vspace{0.1cm} \\
\textbf{Theorem \setword{3}{theorem 3}.} Fix $\lambda$ and $\gamma$ arbitrarily. Let $\left( U^{\left( k \right)},V^{\left( k \right)} \right)$ be the $k$-th iteration generated by lines 2-10 of Algorithm \ref{alg} for solving Problem 5 and $\mathfrak{C}$ the set of KKT points of Problem 5. Under Rules 1 to 2 and Assumption 2, we have $
\lim_{k\rightarrow \infty} \,\,\mathrm{dist}\left( \left( U^{\left( k \right)},V^{\left( k \right)} \right) , \mathfrak{C} \right)=0$. \vspace{0.1cm} 

Assumption 2 appears to be unguaranteed  at first glance in Problem 5, but, in practice, we can artificially anchor one distance sensor without any loss of generality, such that the sequence of variables corresponding to this sensor is constant, and Assumption 2 naturally holds as long as $\mathcal{G}$ is connected.

\textbf{Refinement and extracting estimates.}  Solving Problem 5 with lines 2-10 of Algorithm \ref{alg} results in a high-dimensional realization of $\boldsymbol{p}$, i.e., $U$ (or $V$), in $\mathbb{R}^r$. We propose a refinement step by setting $r=d$ in Problem 5, as line 11 in Algorithm \ref{alg}, which iterates from the previous solution $Q^TU$ (or $Q^TV$). It is justified by the fact that when $r$ is chosen to be exactly the dimension $d$, $Q^TU$ (or $Q^TV$) itself can be considered a realization in $\mathbb{R} ^d$ and thus an (approximate) stationary point of Problem 2 according to Theorem 3. Then, we can recover the state estimates according to Proposition~2.  \vspace{-0.1cm}
   
\subsection{Convergence Guarantees}
\vspace{-0.1cm}
\label{sec:6C}
In this subsection, we show how applying Theorem 1 can lead to the claimed convergence properties Theorems 2 and 3. 

\textbf{Boundness of the seqences.} Theorem 1 assumes the boundness of sequences. We start by declaring that this holds under Assumption 2. \vspace{0.1cm} \\
\textbf{Lemma 2 \textnormal{(Boundness of level sets)}.} Under Assumption 2, the level sets $\mathcal{H}(\bar{\boldsymbol{x}}) =\left\{ \boldsymbol{x}\in \mathcal{X}: f\left(\boldsymbol{x} \right) \leq f\left( \bar{\boldsymbol{x}} \right) , \bar{\boldsymbol{x}}\in \mathcal{X} \right\}$, for a bounded $\bar{\boldsymbol{x}}$, in Problem 4 and Problem 5 are bounded. \vspace{0.1cm} 

The proof of Lemma 2 is provided in Appendix \ref{appendix IV}. Remark~2.2 in \cite{xu2013block} indicates that the boundness of the level set sufficiently guarantees the boundness of sequences generated by BCD. So, the boundness assumption is satisfied.

%
\textbf{Equivalence between Nash equilibriums and KKT points.} Theorem 1 points out the global convergence to a Nash point. We improve this result to critical points for our proposed methods. In particular, the following Lemma reveals the equivalence between Nash equilibriums and KKT points under Assumption 2, and the proof is given in Appendix \ref{appendix IV}.\vspace{0.1cm} \\
\textbf{Lemma 3 \textnormal{(Equivalence between Nash equilibriums and KKT points over a Cartesian product set)}.} Consider an optimization problem (\ref{eq:opt}) given a block division  $\{\varPhi _q\}_{q=1,...,p}$ such that \vspace{-0.15cm} 
\begin{equation}
	\label{eq:Cartesian}
	\mathcal{X} =\mathcal{X} _1\times ...\times \mathcal{X} _p.
	\vspace{-0.15cm}
\end{equation}
Let $f\left( \cdot \right)$ be continuously differentiable. The Nash equilibrium of the problem (\ref{eq:opt}) also satisfies the KKT conditions if the constraints are continuously differentiable and under some qualifications (e.g., the LICQ \cite[Chapter 12.2]{nocedal1999numerical}).  \vspace{0.1cm} 

We note that $\mathcal{X}$ is not a Cartesian product (\ref{eq:Cartesian}) with constraint (\ref{eq:edge constraint}) or the $U$-$V$-coupled constraints under Rule 2 that is necessary for Proposition 4. Fortunately, as long as these constraints are replaced with some surrogate function $s(\boldsymbol{x})$ on the objective function, e.g., as fulfilled by Assumption 3 for ESDP-BCD and the penalty terms in (\ref{eq:cost 2}) for BM-BCD, we can still obtain a Cartesian product structure under Rule 1.  

\textbf{Application of Theorem 1 under specialized block division.} According to Lemma 1, under Rules 1 and 2, the corresponding objective functions of the subproblems of Problem 4 and Problem 5 are strongly convex. That is, Assumption 1 that is required by Theorem 1 is satisfied. Combined with the boundness of sequences (Lemma 2) and equivalence between Nash equilibriums and KKT points (Lemma 3), a direct reference to Theorem 1 proves that BCD converges to critical points of our proposed optimization models. For convex problems, such as Problem 4, the critical point is also one of the global minimizers. Therefore, Theorem 2 and Theorem 3 hold. \vspace{-0.05cm}


\begin{figure*}[htbp]
	\centering
	\raisebox{-0.3cm}{
	\begin{minipage}[b]{0.23\textwidth}
		\subfigure[]{
		\centering
		\includegraphics[width=\columnwidth]{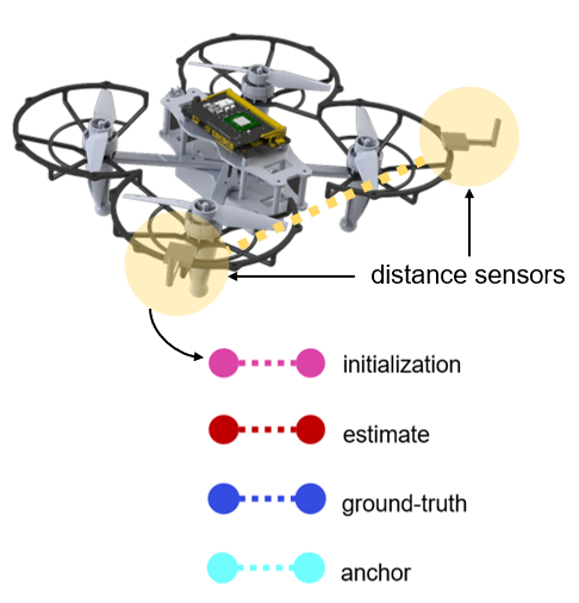}
		\renewcommand\thefigure{\arabic{figure}(a)}
		\label{fig:big_image}
	}
	\end{minipage}
	}
	\hfill
	\begin{minipage}[b]{0.73\textwidth}
		\centering
		\begin{minipage}{0.5\textwidth}
			\subfigure[]{
			\centering
			\includegraphics[width=\textwidth]{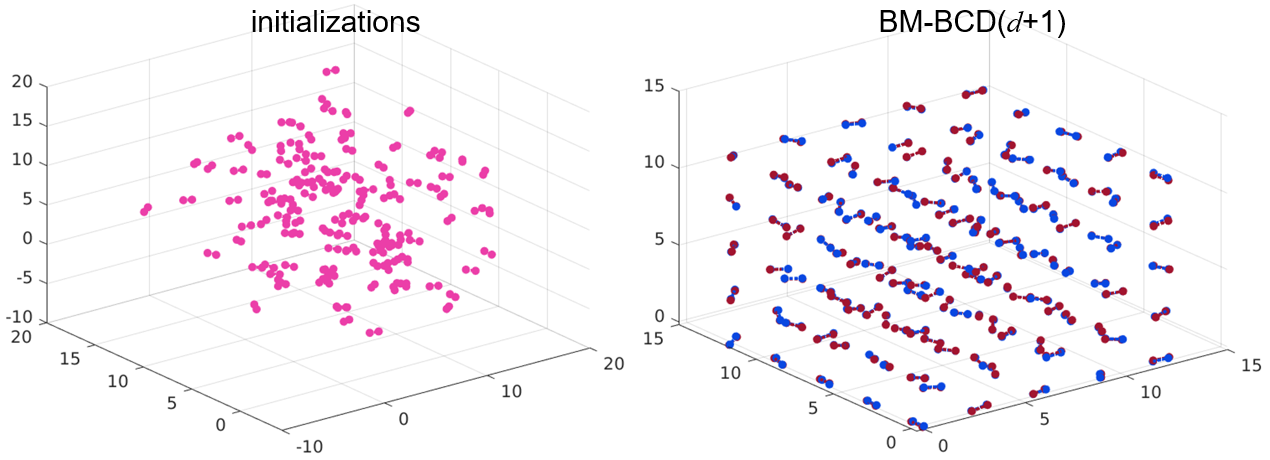}
			\label{fig:simulation3D (a)}
		}
		\end{minipage}\hfill 
		\begin{minipage}{0.5\textwidth}
			\subfigure[]{
			\centering
			\includegraphics[width=\textwidth]{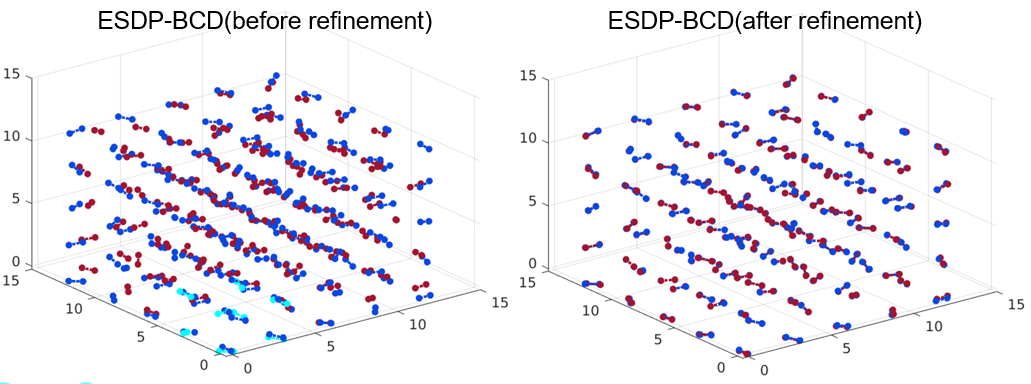}
			\label{fig:simulation3D (b)}
		}
		\end{minipage}
		\begin{minipage}{0.5\textwidth}
			\vspace{-0.2cm}
			\subfigure[]{
			\centering
			\includegraphics[width=\textwidth]{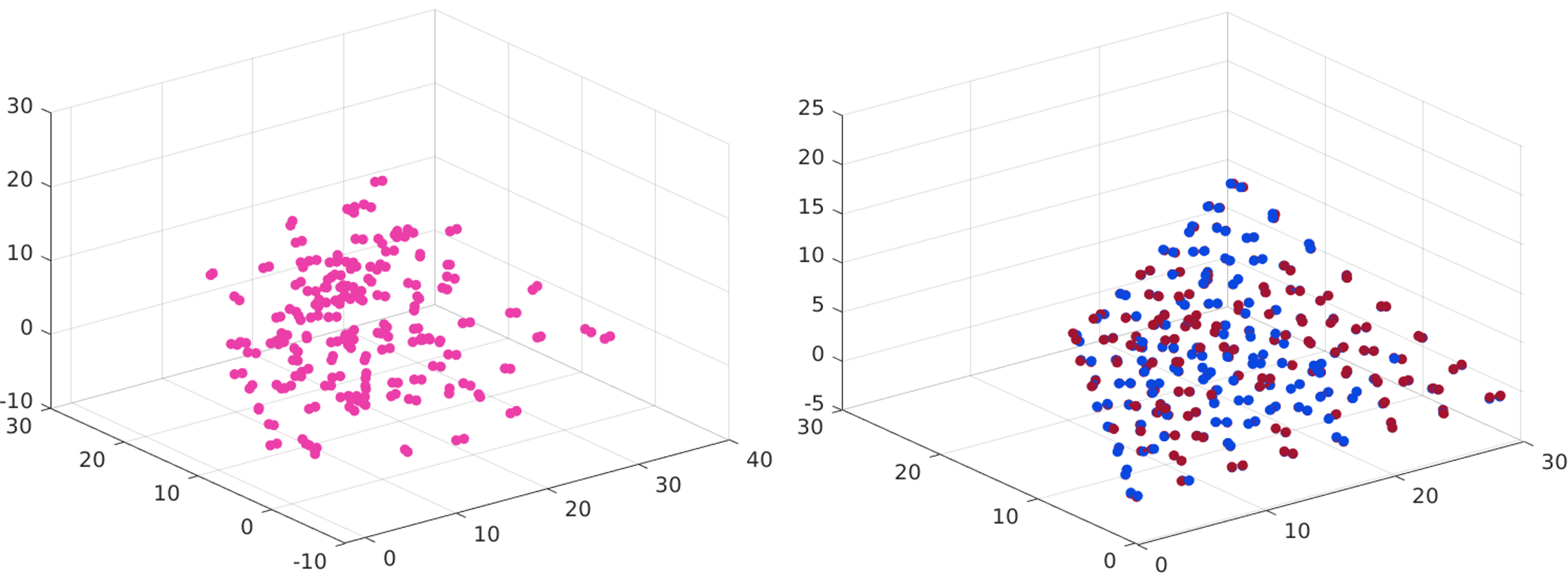}
			\label{fig:simulation3D (c)}
		}
		\end{minipage}\hfill
		\begin{minipage}{0.5\textwidth}
			\vspace{-0.2cm}
			\subfigure[]{
			\centering
			\includegraphics[width=\textwidth]{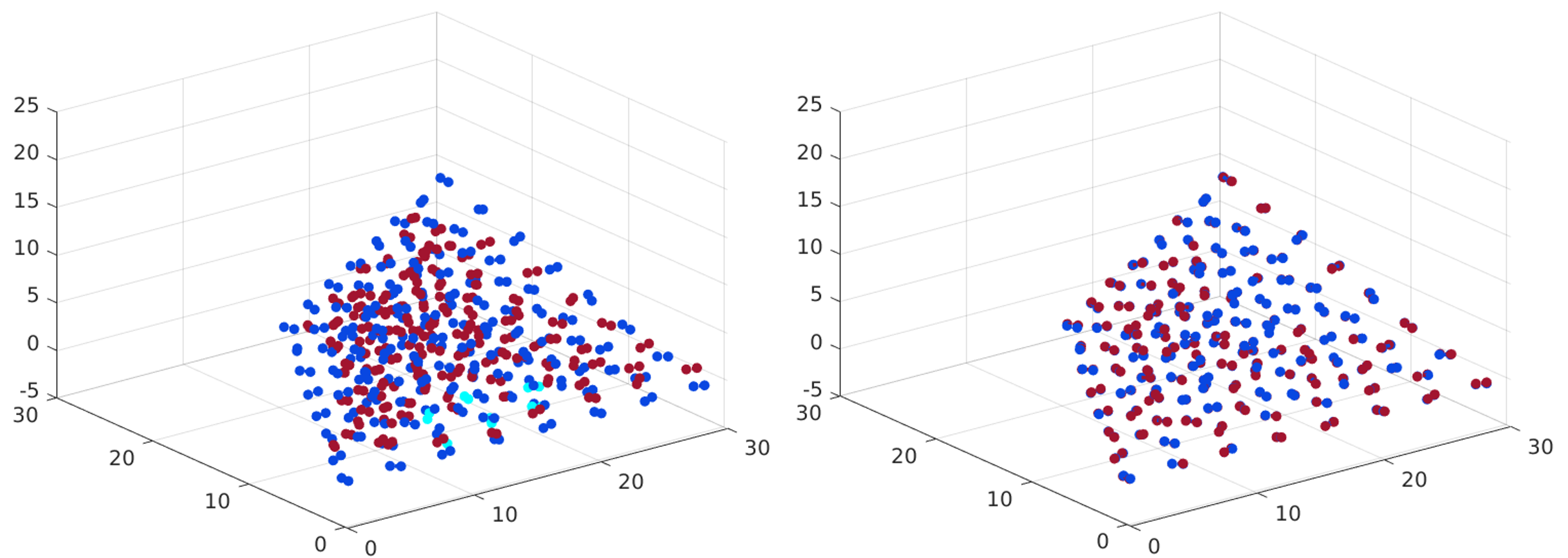}
			\label{fig:simulation3D (d)}
		}
		\end{minipage}
	\end{minipage}
	
	\captionsetup{labelformat=simple}
	\captionsetup{font=footnotesize}
	\caption{ \textbf{Visualization of the problems and results of the proposed methods.} (a) Illustration of a physical agent corresponding to the sensor configuration in Section \ref{sec:7A}: an agent can be graphically abstracted as two connected dots together with a dashed line connecting them, where the solid dots represent the two distance sensors on the agent.  (b) Initialization and results of BM-BCD($d$+1) in \textbf{cube}. (c) Results before (left) and after (right) refinement of ESDP-BCD in \textbf{cube}. (d) Initialization and results of BM-BCD($d$+1) in \textbf{pyramid}. (e) Results before (left) and after (right) refinement of ESDP-BCD in \textbf{pyramid}.}
	\vspace{-0.4cm}
	\label{fig:combination}
\end{figure*}

\section{Experiments}
\vspace{-0.1cm}
\label{sec:7}
In this section, we assess the performance of our proposed methods compared to alternative methods under various system setups.  We perform all experiments in MATLAB running on a Linux laptop with the Intel i5-11400H.

\textbf{Metrics.} We use the \mbox{$\left( \varDelta +1 \right)$-coloring} algorithm \cite{barenboim2013distributed}, where $\varDelta$ is the maximum degree of the dependency graph, for parallelization by default, so that the number of colors can be controlled constant as systems scaling up with a constant $\varDelta$. We evaluate the total time to solve subproblems in parallel, abbreviated as PT (Parallel Time). We also show the total time  to solve each subproblem serially, abbreviated as ST (Serial Time). To measure the accuracy of the relative state estimation, we use the average of the root-mean-squared-error (RMSE) for relative translations in body reference systems for all evaluated methods.

\textbf{Choice of parameters in the algorithms.} The parameter $r$ in BM-BCD affects the results to a great extent. A numerical experiment demonstrates the performance of the algorithm under various $r$ in Fig. \ref{fig:r}, Appendix \ref{appendix V}. From the results, we can see that larger $r$ results in better robustness to the initializations at the cost of taking more iterations. In the following experiments we take $r=d+1$ and $r=d$ (i.e., refinement-only) for comparison. This choice is adequate to highlight the algorithm's features and ensure its efficiency.

The termination condition $\epsilon$ in Algorithm 1 for BM-BCD is uniformly taken as $5\times 10^{-4}$. For ESDP-BCD and SOCP-BCD, $\epsilon$ is taken as $1\times10^{-3}$ in Section \ref{sec:7B} and $5\times10^{-3}$ in Section \ref{sec:7A} and Section \ref{sec:7C}. We sometimes choose a loose termination condition for the latter since they do not provide highly accurate estimates and, therefore, do not need to be 'fully convergent' before refinement.    \vspace{-0.1cm}

\begin{figure} \centering 
	\setcounter{subfigure}{0} 
	\hspace{-0.3cm}
	\subfigure[]{
		{\includegraphics[width=0.468\columnwidth]{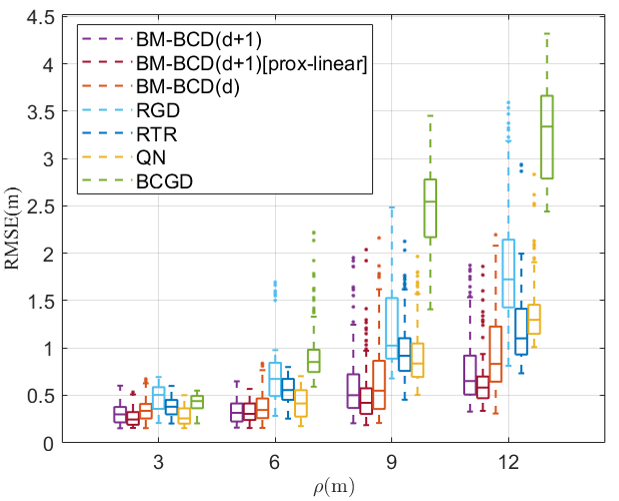}
			\label{fig:NLP precision}}
	} 
	\hspace{-0.3cm}     
	\subfigure[]{  
		{\includegraphics[width=0.469\columnwidth]{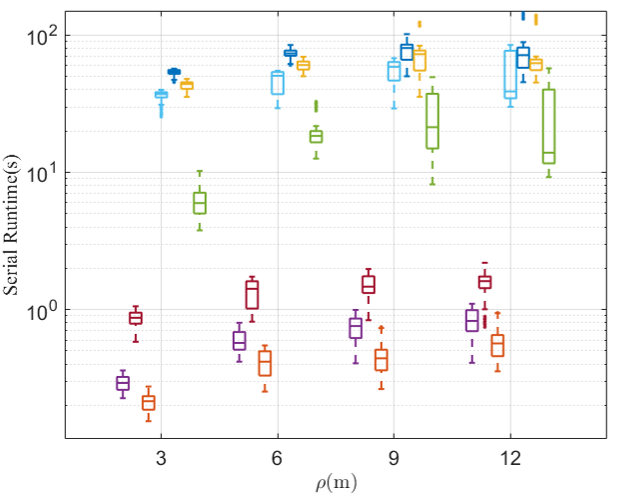}
			\label{fig:NLP runtime}}     
	}  \\  
	\captionsetup{labelformat=simple}
	\captionsetup{font=footnotesize}
	\vspace{-0.2cm}  
	\caption{\textbf{Estimation precision and serial runtime of the NLP methods.} The evaluation is performed in the 125 randomly-distributed-agents scenario. Statistics are computed under 100 times of data generation.}   
	\vspace{-0.4cm}
	\label{fig:NLP}
\end{figure}

\vspace{-0.05cm}
\subsection{Distance-Proprioception-based Systems}
\vspace{-0.1cm}
\label{sec:7A}
In this subsection, we intend to validate the efficacy and efficiency of the methods on problems filled with spurious minima.  We consider 3D systems that consist of agents with two distance sensors and an IMU on each.  

\textbf{Simulation problems.} We test on three simulation problems where agents respectively 'form' a \textbf{cube} (125 agents; Fig. \ref{fig:simulation3D (a)}), a \textbf{pyramid} (120 agents; Fig. \ref{fig:simulation3D (c)}) and are randomly distributed in $[0,12\mathrm{m}]^3$ (125 agents), with sparse-connected measurement topologies $\mathcal{G}$. The $($maximum degree, minimum degree$)$ of $\mathcal{G}$ in the three problems are $(26, 9)$, $(24, 6)$ and $(18, 18)$, respectively. All distance measurements of an agent occur with the specific number of  nearest neighboring  agents. The ground-truth distance between adjacent agents in the first two problems are $3\mathrm{m}$ and $4\mathrm{m}$, respectively.  The default distance noise level is $\sigma=0.1\mathrm{m}$, and the errors of each attitude angle measurement are uniformly distributed in $[-1.5\deg, 1.5\deg]$, aligned with those observed in hardware experiments \cite{shalaby2021relative}. The positions of distance sensors are uniformly set as $\bar{\nu}_{i}^{0}=\left[ 0;0.35\mathrm{m};0 \right]$ and $\bar{\nu}_{i}^{1}=\left[ 0;-0.35\mathrm{m};0 \right]$. We randomly generate the initial guess of each agent's location on a ball with radius $\rho$, and  the ground-truth as its center. For convex-relaxation methods, we set 8 and 6 \emph{imperfect} anchor agents in the \textbf{cube} and \textbf{pyramid} problems, respectively, each of which measures 15 other agents. See Fig. \ref{fig:simulation3D (b)} and Fig. \ref{fig:simulation3D (d)}.

\subsubsection{Comparison of local search methods}
 \textbf{Benchmark.} When using local search methods standalone, we compare the proposed method BM-BCD with (i) \emph{Riemannian Gradient Descent} (RGD), a  first-order algorithm for solving Problem 1 on smooth Riemannian manifolds based on a centralized backtracking line-search, previously used to solve a similar problem in \cite{cossette2022optimal}, (ii) \emph{Riemannian Trust-Region} (RTR), a numerical robust centralized second-order algorithm for solving Problem 1 on smooth Riemannian manifolds, (iii) \emph{Quasi-Newton} (QN), a standard centralized quadratic first-order algorithm for solving Problem 5 with a fixed $r = d$, (iv) \emph{Block Coordinate Gradient Descent} (BCGD), a BCD-type algorithm which replaces all the block updating steps in BM-BCD with backtracking line-search-based gradient descent with $r = d+1$. The Riemannian optimization methods are implemented in the Manopt Toolbox \cite{boumal2014manopt}, where Hessian is approximated using finite differences. QN is implemented in the MATLAB Optimization Toolbox.
 
We evaluate three variants of BM-BCD. BM-BCD($d$) and BM-BCD($d$+1) use update (\ref{eq:block update problem}), where the parameter $r$ in Problem 5 is set as $d$ and $d$+1, respectively, while BM-BCD($d$+1)[prox-linear] uses update (\ref{eq:prox-linear update}).

\textbf{Results.} Fig. \ref{fig:NLP} shows the performance of each method in the distance-proprioception setup. Fig. \ref{fig:simulation3D (a)} and Fig. \ref{fig:simulation3D (c)} visualize the results of BM-BCD($d$+1) with $\rho$=6m and $\rho$=8m, respectively. From the results, we can see that among the local search methods, the BM-BCD family and QN manifest desired precision under mild initialization conditions, whereas the methods based on Riemannian optimization prone to fall into worse local minima than BM-BCD and QN. This may be an inherent disadvantage of Riemannian optimization algorithms, as its non-convex search space may lead to ill-conditioned local geometry and introduce additional shallow local minima. Compared to QN, the computational efficiency provided by the BM-BCD method is irreplaceable, with magnitudes of advantage. {The} method that also iterates fast is BCGD.   However, due to the lack of careful design, BCGD \emph{fails} to converge to desired termination conditions sometimes. 

One of the promising results {shown} in Fig. \ref{fig:NLP precision} is that the BM-BCD family exhibits robustness  to initial conditions, which is unmatched by all other methods. Among the BM-BCD family, BM-BCD($d$+1)[prox-linear] exhibits the most stable performance, probably because the local linearization in the update (\ref{eq:prox-linear update}) helps the method to escape {certain local minimas} \cite{zeng2019global,xu2013block}, at the cost of taking more iterations. 

\subsubsection{Comparison of convex relaxation methods}
\label{sec:convex relaxation exp}
\begin{figure} \centering 
	\setcounter{subfigure}{0} 
	\hspace{-0.2cm}
	\subfigure[]{
		\raisebox{0.25ex}
		{\includegraphics[width=0.46\columnwidth]{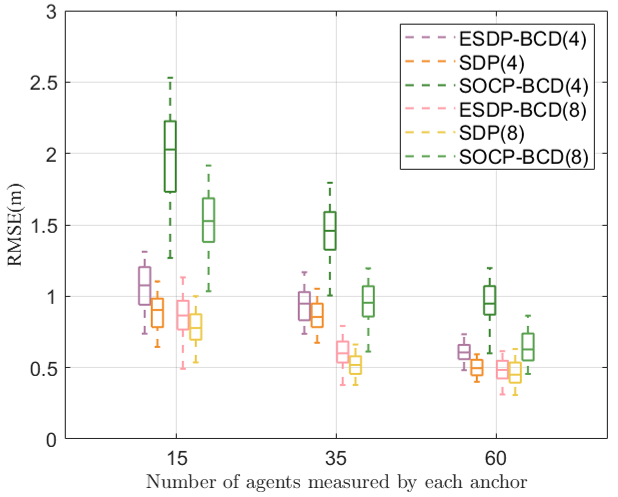}
			\label{fig:rate precision}}
	} 
	\hspace{-0.5cm}     
	\subfigure[]{  
		{\includegraphics[width=0.478\columnwidth]{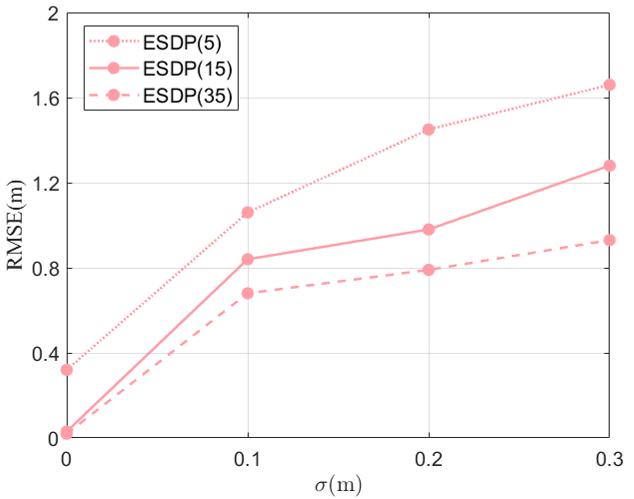}
			\label{fig:noise precision}}     
	} \vspace{-0.3cm} \\  
	\captionsetup{labelformat=simple}
	\captionsetup{font=footnotesize}  
	\caption{\textbf{Convex relaxation-based methods under various numbers of anchor measurements.}  The evaluation is performed in the 125 randomly-distributed-agents scenario. Statistics are computed under 100 times of data generation for ESDP-BCD and SOCP-BCD and 30 times for SDP. (a) The benchmark results under various numbers of anchor measurements, where the number marked in the legend represents the number of anchors. (b) The results of different numbers of measurements per anchor with 6 anchors under various noise levels, where the number marked in legend represents the number of anchor measurements.}   
	\vspace{-0.4cm}
	\label{fig:anchor rate}
\end{figure}

\textbf{Benchmark.} The convex relaxation-based ESDP-BCD method is compared with (i) \emph{Semidefinite Programming} (SDP), a direct rank-relaxed version of Problem 3, (ii) \emph{Second-Order Cone Programming with Block Coordinate Descent} (SOCP-BCD), a second-order cone relaxation \cite[Eq. (6)]{tseng2007second} of Problem 3 with anchors distributed at the boundary of the system\footnote{SOCP-BCD demands strictly on the distribution of anchors in our experiments, so we use an (impractical) approach that lets the anchors be scattered across the boundary to allow SOCP to produce valid estimates.}, which is \emph{structural decomposable} and thus can be solved via a BCD-style algorithm as ESDP-BCD. For a fair comparison, SDP, RE and ESDP-BCD share the same anchor information. SDP and SOCP-BCD also use BM-BCD as refinement in Table \ref{table1}, while the shown errors in the experiment of Fig. \ref{fig:anchor rate} are all calculated \emph{before refinements} and in the common reference system instead of body reference systems. We solve convex programming in the above methods using the interior point method implemented in the MOSEK solver \cite{aps2019mosek}. 

\textbf{Results.} Convex relaxation-based methods result in relatively consistent solutions on a given setup due to its robustness to initial guesses. As can be seen in Table \ref{table1}, SDP results in the minimum RMSE before refinement compared to other methods and ESDP-BCD is the second, and the results of the two methods are \emph{close}. After the refinement stage, the two methods behave \emph{similarly} in terms of precision, although ESDP is a further relaxation of SDP. This is due to the robustness to initialization provided by BM-BCD. SOCP-BCD results in a highly suboptimal solution due to too little anchor information. In addition, due to the ease with which ESDP-BCD and SOCP-BCD can take advantage of initial guesses and their decomposable structures, these methods exhibit higher computational efficiency than SDP in such large-size problems. We postpone a fairer comparison of scalability to the distance-only setup in the following subsection where ESDP-BCD is randomly initialized.

We also show the behavior of the convex relaxation methods under different anchor and anchor measurement configurations in Fig. \ref{fig:anchor rate}.  It can be seen that increasing the number of anchors and anchor measurements can both help the convex relaxation-based methods perform better in accuracy, where SOCP-BCD is most sensitive to changes of such conditions. \vspace{-0.2cm}

\begin{table}[]
	\vspace{0.2cm}
	\setlength{\tabcolsep}{2pt}
	\renewcommand\arraystretch{1.1}
	\centering
	\hspace{-0cm}\begin{tabular}{ccccc}
		\hline	Problem   & Method & \footnotesize{PT(s)} & \footnotesize{ST(s)} & \footnotesize{$\rm RMSE$}(m)   \\ \hline
		\footnotesize{\textbf{cube}} 
		& \textcolor{magenta}{ESDP-BCD} & \textbf{7.12} & 33.7 & \textbf{0.23} (1.28)  \\
		 & SDP & ------ & $>$100 & \textbf{0.23} (0.86)  \\
		& SOCP-BCD & 8.84 & 38.9 & 0.36 (2.08)  \\
		\hline
		\footnotesize{\textbf{pyramid}} 
		& \textcolor{magenta}{ESDP-BCD} & \textbf{6.23} & 35.8 & \textbf{0.40} (1.54)  \\
		 & SDP & ------ & 93.9 & \textbf{0.38} (1.32)  \\
		& SOCP-BCD & 7.64 & 44.2 & 0.50 (2.53)  \\
		\hline
		\vspace{-0.2cm}
	\end{tabular}
	\captionsetup{labelformat=simple}
	\captionsetup{font=footnotesize} 
	\caption{\textbf{Numerical results of the proposed and alternative methods.} The methods labeled using \textcolor{magenta}{magenta} are proposed in this paper. The accuracy metric RMSE is presented as 'RMSE in body reference systems after refinement (RMSE in the common reference system before refinement)'. Statistics are computed as the average results under 100 times of data generation for each problem and each method except SDP. SDP is tested 30 times for data generation on each problem. }
	\vspace{-0.3cm}
	\label{table1}
\end{table}

\begin{table}[ht]
	\setlength{\tabcolsep}{2pt}
	\renewcommand\arraystretch{1.1}
	\centering
	\hspace{-0cm}\begin{tabular}{cccc}
		\hline	Problem   & Method &  \footnotesize{$\rm RMSE$}(m)   \\ \hline
		\textbf{cube} & \textcolor{magenta}{ESDP-BCD} & \textbf{0.48}$\pm$0.11 (0.57$\pm$0.13)  \\
		& ESDP-BCD (refined with RTR) & 0.52$\pm$0.13 (0.57$\pm$0.13)  \\
		& RE & \textbf{0.45}$\pm$0.08 (0.43$\pm$0.11)  \\
		& SDP & \textbf{0.45}$\pm$0.09 (0.53$\pm$0.15)  \\
		& SDP (refined with RTR) & 0.49$\pm$0.10 (0.53$\pm$0.15)  \\
		& SOCP-BCD & 0.79$\pm$0.31 (1.24$\pm$0.29)   \\
		\hline
		\textbf{hexagon} & \textcolor{magenta}{ESDP-BCD} & \textbf{0.32}$\pm$0.09 (1.08$\pm$0.23)  \\
		& ESDP-BCD (refined with RTR) & 0.40$\pm$0.11 (1.08$\pm$0.23)  \\
		& RE & \textbf{0.35}$\pm$0.07 (0.75$\pm$0.15)  \\
		& SDP & \textbf{0.31}$\pm$0.08 (0.86$\pm$0.17)  \\
		& SDP (refined with RTR) & 0.37$\pm$0.10 (0.86$\pm$0.17)  \\
		& SOCP-BCD & 0.82$\pm$0.36 (1.82$\pm$0.43)   \\
		\hline
		\vspace{-0.3cm}
	\end{tabular}
	\captionsetup{labelformat=simple}
	\captionsetup{font=footnotesize} 
	\caption{\textbf{Numerical results of the proposed and alternatives methods.} The methods labeled using \textcolor{magenta}{magenta} are proposed in this paper. The accuracy metric RMSE is presented as 'RMSE in body reference systems after refinement (RMSE in the common reference system before refinement)'. The results of RE are obtained from 20 times of data generation, and other experiment methods and setups are consistent with those of Fig. \ref{fig:anchor rate}.  }
	\label{table2}
	\vspace{-0.5cm}
\end{table} 

\subsection{Distance-only Systems}
\vspace{-0.1cm}
\label{sec:7B}
In this subsection, we consider the distance-only setup in 2D and 3D problems, where each agent is equipped with three distance sensors for 3D problems and two for 2D problems. 

\textbf{Simulation problems.} We test on three simulation problems where agents respectively 'form' a 3D \textbf{cube} and a 2D \textbf{hexagon} (217 agents; Fig. \ref{fig:hexagon} in Appendix \ref{appendix V}). The $($maximum degree, minimum degree$)$ of $\mathcal{G}$ in the \textbf{hexagon} problem is $(12, 5)$, respectively. The positions of sensors on each agent are uniformly set as $\bar{\nu}_{i}^{0}=\left[ 0;0.35\mathrm{m}\right]$ and $\bar{\nu}_{i}^{1}=\left[ 0;-0.35\mathrm{m}\right] $ for $d=2$. We set 4 anchor agents in the \textbf{hexagon} problem, each of which measures other 15 agents. The initial guess of each agent's state is randomly distributed in a range for a relatively fair comparison of scalability. The rest of the settings are consistent with the previous subsection.
 
\textbf{Benchmark.} In the distance-only setup, we evaluate the convex relaxation-then-refinement pipeline and introduce \emph{Riemmannian Elevator} (RE) \cite{halsted2022riemannian}, a state-of-the-art method for centralized distance-only localization in the term of solution optimality, as an additional benchmark.  RE uses RTR for refinement in our experiment, consistent with \cite{halsted2022riemannian}. We also use RTR to ablate the refinement algorithm BM-BCD in ESDP-BCD and SDP, as remarked in Table \ref{table2}. The remaining setup of benchmark methods is consistent with that of \ref{sec:convex relaxation exp}.

\textbf{Results.} It can be seen from Table \ref{table2} that all ESDP-BCD, RE, and SDP can provide informed initializations for the refinement stage. The relaxation used in RE results in the best accuracy among the methods, consistent with the previous study's results \cite{halsted2022riemannian}.  This is because the algorithmic design of RE creates a restriction on the rank of the solution. In contrast, SDP and ESDP-BCD use the standard interior-point method solver, leading to the max-rank solution in their solution sets \cite{mehrotra1992implementation}. However, with the help of BM-BCD, the final estimates of ESDP-BCD and SDP can still be better or quite precise than RE, which cannot be achieved with RTR. 

\begin{figure} \centering 
	\setcounter{subfigure}{0} 
	\hspace{-0.2cm}
	\subfigure[]{
		{\includegraphics[width=0.47\columnwidth]{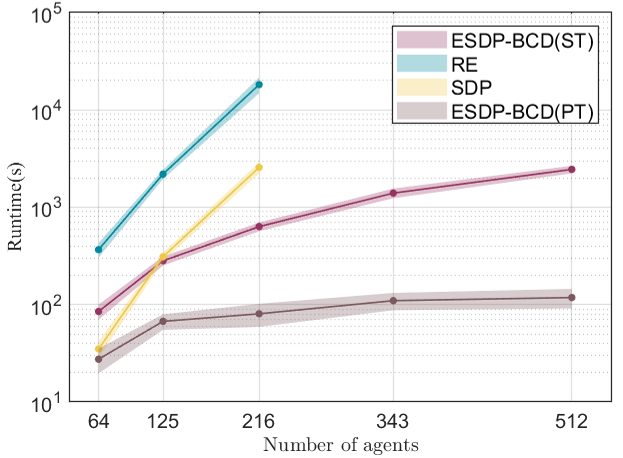}
			\label{fig:scalability}}
	}   
	\hspace{-0.4cm}
	\subfigure[]{
		\raisebox{0.1ex}
		{\includegraphics[width=0.466\columnwidth]{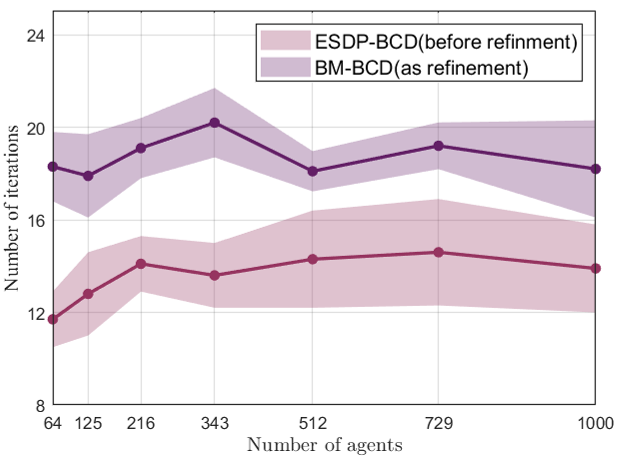}
			\label{fig:iterations}}     
	} \vspace{-0.2cm} \\  
	\captionsetup{labelformat=simple}
	\captionsetup{font=footnotesize}  
	\caption{\textbf{Scalability of the methods under various number of agents.} The evaluation is performed in a system with $l^3$ agents, where $l$ is a positive integer, and the other parameters are consistent with the \textbf{cube} problem. The initial guess of each agent's state is randomly distributed in the cube for a relatively fair comparison. Statistics are computed under 20 times of data generation for ESDP-BCD and SDP, 20 times for RE at less than 216 agents, and 4 times at 216 agents. When the number of agents exceeds 216, the SDP and RE methods cause the computer to be out-of-memory.}   
	\vspace{-0.6cm}
	\label{fig:scalability evaluation}
\end{figure}

Of particular importance, as shown in Fig. \ref{fig:scalability evaluation}, ESDP-BCD exhibits scalability that SDP and RE \emph{cannot} match due to its decomposable structure and the adoption of the BCD algorithmic framework. In particular, the number of iterations in both stages of ESDP-BCD remains approximately \emph{constant} as the system is scaled up, as highlighted in Fig. \ref{fig:iterations}. \vspace{-0.2cm}

\begin{figure}[b] \centering 
	\vspace{-0.3cm}
	\setcounter{subfigure}{0} 
	\hspace{-0.3cm}
	\subfigure[]{
		{\includegraphics[width=0.35\columnwidth]{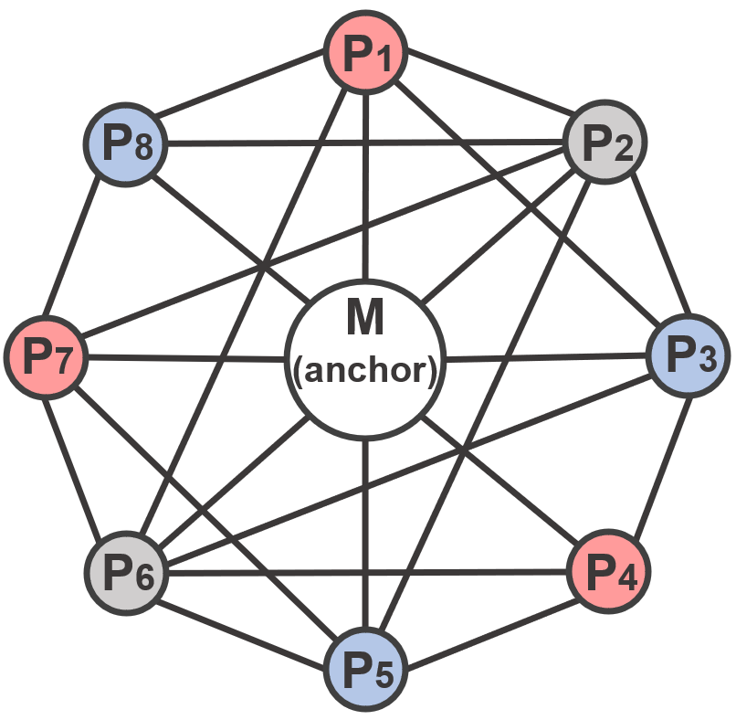}
			\label{fig:topology}}
	} 
	\hspace{-0.3cm}     
	\subfigure[]{  
		{\includegraphics[width=0.45\columnwidth]{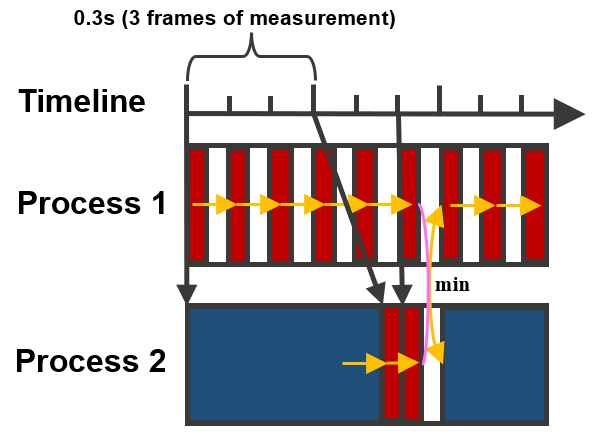}
			\label{fig:pipeline}}     
	} \vspace{-0.2cm} \\  
	\captionsetup{labelformat=simple}
	\captionsetup{font=footnotesize}  
	\caption{\textbf{Illustration of the mothership-parasite system.} (a) The measurement topology and its coloring scheme, where 'M' indicates the mothership and P$_i$ indicates the $i$-th parasite. (b) The pipeline of the combination of BM-BCD and ESDP-BCD. The figure shows how each of the two processes works and coordinates with each other, where the red and blue rectangles indicate the execution of BM-BCD and ESDP-BCD, respectively, the yellow arrows indicate the estimation results of one frame offering initialization for the subsequent optimization, and the pink curves indicate the fusion of the estimation results from the two processes (by taking the one with the lower objective function value).}   
	\label{fig:mothership}
\end{figure}

\begin{figure*} \centering
	\hspace{-0.2cm}\subfigure[] {
		  
		\raisebox{0.01cm}  
		{\includegraphics[width=0.8\columnwidth]{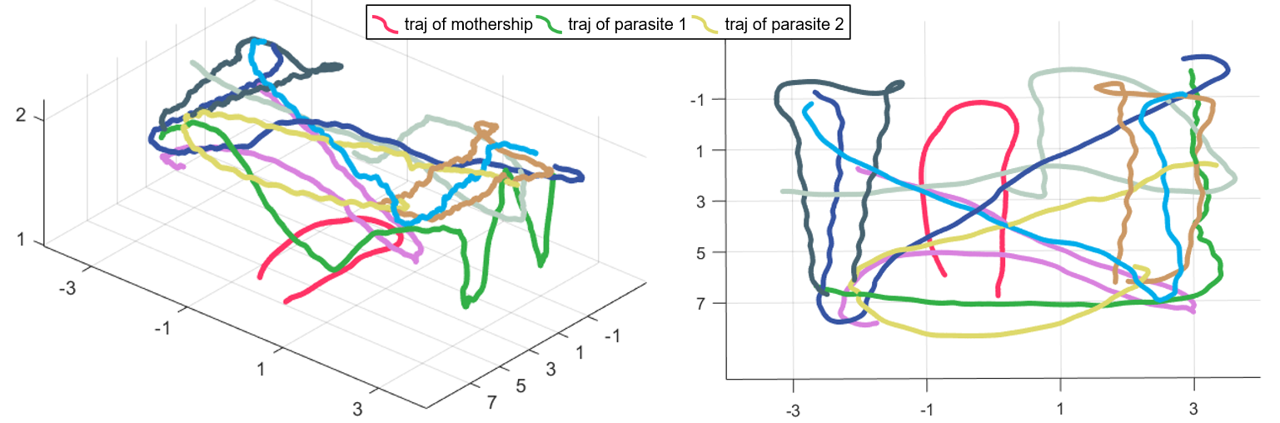} 
			\label{fig:trajectory} }
	}      
	\hspace{-0.3cm}\subfigure[] { 
		\label{fig:result parasite1}     
		\includegraphics[width=0.8\columnwidth]{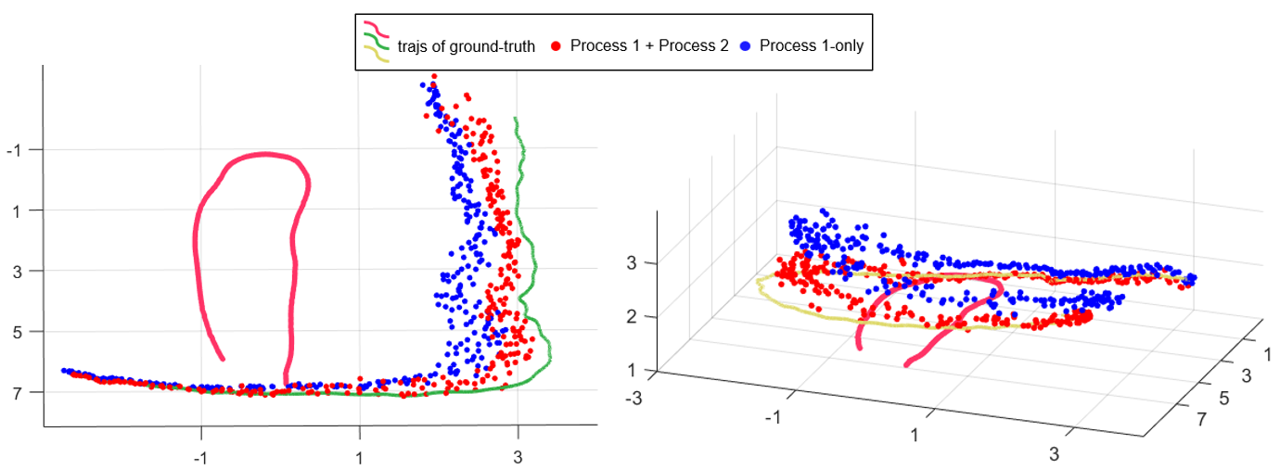}     
	} 
	\hspace{-0.3cm}\subfigure[] {  
		\label{fig:result eta}  
		\includegraphics[width=0.4\columnwidth]{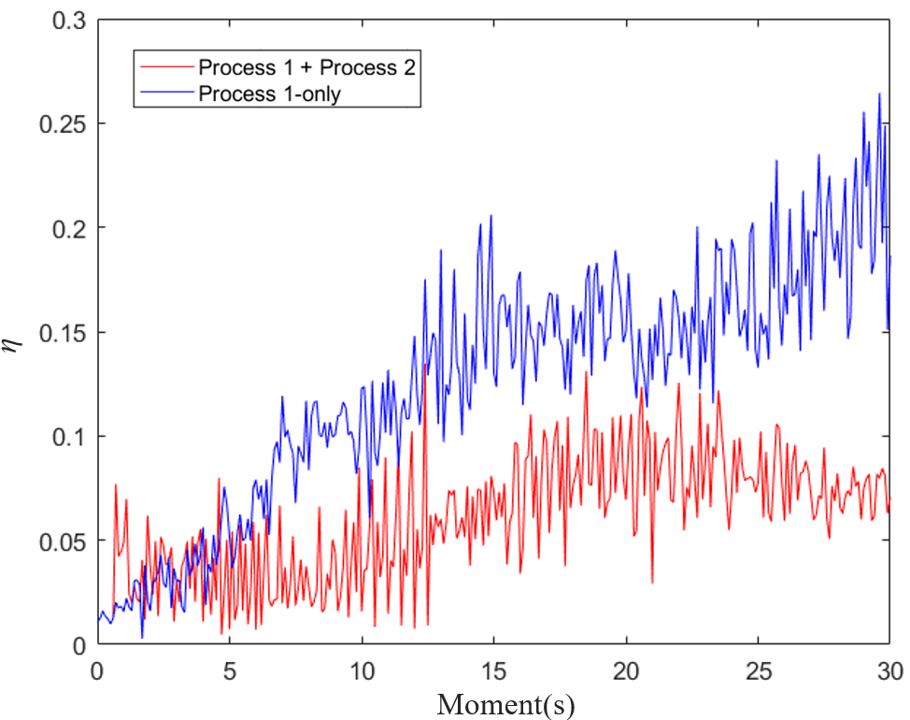}     
	}
	\captionsetup{labelformat=simple}
	\captionsetup{font=footnotesize} 
	\vspace{-0.2cm}
	\caption{ \textbf{Ground-truth and estimates from the two methods on the mothership-parasite system.} (a) The 3D view (left) and XY plane view (right) of the ground-truth trajectories of each agent.  (b) Estimation results of parasite 1 (left) and parasite 2 (right) from 'Process 1 + Process 2' (red) and 'Process 1-only' (blue) in the body frame of mothership at each moment. (c) The relative estimation error of the parasite 2 in the body frame of mothership at each moment.}
	\label{fig:trajectory result}
	\vspace{-0.4cm}      
\end{figure*}

\subsection{A Heterogeneous System in a Continuous-Time Scenario}
\label{sec:7C}
One of the classic applications of relative state estimation is the \emph{mothership-parasite system} \cite{leonard2016autonomous,alcocer2006underwater}. In such a system, the mothership is designed to allow for large sensor baselines and an assembly of more sensors that \emph{serve as natural anchors}. In contrast, the parasite supports only relatively small sensor baselines.  We show that combining BM-BCD and ESDP-BCD, i.e., an algorithmic redundancy design, leads to more consistent estimation over time. 

\textbf{Real-world platform setup.} We briefly describe the experimental platform and postpone the details of in Appendix \ref{appendix VI}, which explains how we collect data and calibrate the bias of the UWB distance sensors \cite{fishberg2022multi,fishberg2023murp}. The distance measurements after smoothing are sampled at 10 Hz. The system consists of one mothership and 8 parasites. There are 4 distance sensors on the mothership with the furthest interval distance around $2.3$m. Parasites are equipped with two distance sensors on each with baselines around $0.7$m. Each agent is equipped with an IMU. The ground-truth trajectories of different agents are illustrated in Fig. \ref{fig:trajectory}, each of which is represented by a separate color.  The measurement topology between agents and the coloring scheme are shown in Fig. \ref{fig:topology}. 

\textbf{Methods.} BM-BCD is combined with ESDP-BCD in a flow illustrated in Fig. \ref{fig:pipeline}. In the system of this section, the BM-BCD($d$+1) can operate up to 50 Hz, while the ESDP-BCD can easily operate up to 2 Hz, taking sensors on the mothership as anchors. The method of combination is described below:
(a) On each frame of measurements, Process 1 immediately performs online inference using BM-BCD($d$+1) initialized with the estimate from the last measurement; (b) Process 2 reads the latest distance measurement and infers using ESDP-BCD with the latest estimate as the initialization. Since ESDP-BCD cannot infer at the similar high frequency as BM-BCD, the results of ESDP-BCD may exhibit a latency. To handle this latency, Process 2 samples the measurements generated during the computation of ESDP-BCD in a 0.3s interval and perform BM-BCD($d$+1) with the results of ESDP-BCD as the initialization until the latest frame of measurements have been used. The solution from the two processes with a lower objective function value serves as the final estimate for that frame. Process 1 also runs standalone for comparison.

\textbf{Results.} We show the results of employing both pipelines, 'Process 1-only' or 'Process 1 + Process 2', in Fig. \ref{fig:trajectory}. Since the absolute precision of estimates is related to the relative position between agents, we use a feature number $\eta=\textnormal{RMSE}/\underline{d}_{ij}$, where $\underline{d}_{ij}$ is the ground-truth distance between agent $i$ and $j$, to define a relative precision of the estimation, as used in Fig. \ref{fig:result eta}. It can be seen that $\eta$ can increase over time when only executing Process 1, while executing 'Process 1 + Process 2' results in a more consistent $\eta$. We suppose that the estimation 'drift' of the local search method occurs because the solution of BM-BCD($d$+1) is affected by initializations, i.e., the last estimate in our methods: because of the near-continuous variation of the measurements over time, the landscape of the optimization also varies continuously, such that similar initializations lead to similar local minima in the continuous-time scenario. Therefore, once BM-BCD($d$+1) falls into a shallow local minima, it may be stuck in similar local minima in the subsequent few frames. In contrast, executing ESDP-BCD periodically to 'shuffle' the estimates alleviates such a problem due to the robustness to arbitrary initializations of ESDP-BCD.  \vspace{-0.1cm}

\section{Conclusion and Future Work} 
\vspace{-0.1cm}
\label{sec:conclusion}
This paper presents a systematic algorithmic scheme for the problem of distance-based multi-agent relative state estimation, demonstrating that combining scalability and robustness is possible. Both the scalable global estimation and robust local search stages significantly  contribute to the results. BCD on the multiconvex structures is the key to efficient and safe algorithmic implementation and scalability.

The optimization problem in this paper belongs to the general scope of polynomial optimization problems (POPs) \cite{lasserre2009moments}, which is widely emerging in the field of robotics \cite{teng2023convex,yang2022certifiably,maric2021riemannian}. The current results are limited to formulations including only certain terms in the canonical basis of polynomial, while it can be untrivial to generalize a similar idea of constructing multiconvex  and decomposable models to general POPs. Nevertheless, we wish to bring the scalable/parallelizable solution procedure achieved in this paper to more POP-based robotics problems in the future, especially those that are time-critical such as trajectory optimization \cite{teng2023convex}. 

Moreover, a recent work \cite{wang2024fast} demonstrates that the solution of SOCP with \emph{low-dimensional but massive} constraints, such as (\ref{eq:X>p2})-(\ref{eq:inner sdp}), can be accelerated by magnitudes with a specialized solution algorithm. We hypothesize that ESDP-BCD is potentially largely accelerated by exploiting such a configuration of constraints with a dedicated solver. 

\bibliographystyle{plainnat}
\bibliography{references}
\vspace{-0.3cm}
\appendices

\section{Proof of Proposition 1}
\label{appendix I}
\emph{Proof:} \textbf{The distance-only setup.} In the distance-only setup, the aspect (a) is straightforward due to the rigid body invariance rendered by rotation. For the aspect (b), 
\begin{itemize}
	\item when $\left| \mathcal{B} _i \right|=2$ or all the $p_i^u$ are colinear, it is clear that there must exist infinite $R_i$ that satisfies (\ref{eq:equation}), i.e., the aspect (b) holds.
	\item When $\left| \mathcal{B} _i \right|=3$ and (\ref{eq:norm constraint}) holds, we can apply the result in \cite[Section 5]{horn1987closed} to  register any $\hat{p_{i}^{u}}$, $\hat{p_{i}^{v}}$, $\hat{p_{i}^{w}}$, where $u<v<w$ and $p_{i}^{u}$, $p_{i}^{v}$, $p_{i}^{w}$ are non-collinear, to $\bar{\nu}_{i}^{u}$, $\bar{\nu}_{i}^{v}$, $\bar{\nu}_{i}^{w}$, with a unique solution \cite[Section 2]{horn1987closed}. That is, there must exist a unique $(R_i,t_i)$ that satisfies (\ref{eq:equation}) so that the aspect (b) holds.
	\item When $\left| \mathcal{B} _i \right|>3$ but all the points are coplanar, we can choose any 3 non-collinear points to compute a unique solution $(R_i,t_i)$ as explained above. Then the other points on the plane can be uniquely determined by these 3 points with known distances by (\ref{eq:norm constraint}) from these points to the chosen 3 points. That is, if (\ref{eq:norm constraint}) holds and all the points are coplanar, there must exist a unique $(R_i,t_i)$ that satisfies (\ref{eq:equation}).
\end{itemize}

We note that, if $\left| \mathcal{B} _i \right|>3$ and the points are non-coplanar, we can choose 3 non-collinear points to compute a unique $(R_i,t_i)$, and it can be validated that there exist two sets of other points satisfying (\ref{eq:norm constraint}) with \emph{handedness symmetry} about the plane consisting of the 3 chosen points. Due to the uniqueness of $(R_i,t_i)$, it can be clearly shown that one of the sets must not satisfy condition (\ref{eq:equation}) so that the aspect (b) does not hold. That is, the statement 'coplanar' is necessary.

\textbf{The 4-axis proprioception setup.} Firstly, we can show that 
\begin{equation}
	\label{eq:p=Rv+t2dp=Rdv}
	\begin{split}
		p_{i}^{u}=R_i\bar{\nu}_{i}^{u}+t_i,\forall i\in \mathcal{V} ,u\in \mathcal{B} _i\Leftrightarrow \\ p_{i}^{u}-p_{i}^{v}=R_i\left( \bar{\nu}_{i}^{u}-\bar{\nu}_{i}^{v} \right) ,\forall i\in \mathcal{V} ,u,v\in \mathcal{B} _i.
	\end{split}
\end{equation}

When 4-axis proprioception is available, the right-hand side of (\ref{eq:p=Rv+t2dp=Rdv}) can be specially written as  
\begin{equation}
	\label{eq:dp=Rv}
	\begin{split}
			p_{i}^{u}-p_{i}^{v}=R_i\left( \bar{\nu}_{i}^{u}-\bar{\nu}_{i}^{v} \right) \ \ \ \ \ \ \ \ \ \ \ \ 
			\\ =\left[ \begin{matrix}
					\cos \psi _i&		-\sin \psi _i&		0\\
					\sin \psi _i&		\cos \psi _i&		0\\
					0&		0&		1\\
				\end{matrix} \right]\tilde{R}_{i|y}\hat{R}_{i|x}\left( \bar{\nu}_{i}^{u}-\bar{\nu}_{i}^{v} \right),
		\end{split}
	\vspace{-0.2cm}
\end{equation}
where $\hat{R}_{i|y}$ and $\hat{R}_{i|x}$ are the measured rotation components.

Introducing $
\left[ \bar{\nu}_{i|x};\bar{\nu}_{i|y};\bar{\nu}_{i|z} \right]\coloneqq \hat{R}_{i|y}\hat{R}_{i|x}\left( \bar{\nu}_{i}^{u}-\bar{\nu}_{i}^{v} \right)$ for the sake of concise writing, a straightforward computation shows the following relations from the first two rows of (\ref{eq:dp=Rv}):
\begin{equation}
	\label{eq:sin}
	\sin \hat{\psi}_i={{\left( \bar{\nu}_{i|x}\left( \hat{y}_{i}^{u}-\hat{y}_{i}^{v} \right) -\bar{\nu}_{i|y}\left( \hat{x}_{i}^{u}-\hat{x}_{i}^{v} \right) \right)}/{( {\bar{\nu}_{i|x}}^2+{\bar{\nu}_{i|y}}^2)}},
\end{equation}
\vspace{-0.4cm}
\begin{equation}
	\label{eq:cos}
	\cos \hat{\psi}_i={{\left( \bar{\nu}_{i|x}\left( \hat{x}_{i}^{u}-\hat{x}_{i}^{v} \right) +\bar{\nu}_{i|y}\left( \hat{y}_{i}^{u}-\hat{y}_{i}^{v} \right) \right)}/{( {\bar{\nu}_{i|x}}^2+{\bar{\nu}_{i|y}}^2 )}}, \vspace{-0.1cm}
\end{equation}
both of which  hold for any $i\in \mathcal{V}$ and $u,v\in \mathcal{B} _i$. So far, a one-to-one correspondence between $\{({p}_{i}^{u},{p}_{i}^{v})\}$ and $\{({R}_{i},{t}_{i})\}$ is found, which is used in Proposition 2 to recover states from the sensor coordinates.

Analogously, the last row of (\ref{eq:dp=Rv}) gives that\vspace{-0.2cm}
\begin{equation}
	\label{eq:dz}
		\begin{aligned}		
			z_{i}^{u}-z_{i}^{v}&=\bar{\nu}_{i|z} \\ &=
			\left[ \begin{matrix}
				-\sin \tilde{\phi}_i,\ \cos \tilde{\phi}_i\sin \tilde{\theta}_i,\ \cos \tilde{\phi}_i\cos \tilde{\theta}_i\\
			\end{matrix} \right] \left( \bar{\nu}_{i}^{u}-\bar{\nu}_{j}^{v} \right)\hspace{-0.05cm},
		\end{aligned}
	\vspace{-0.2cm}
\end{equation} 
which appears in (\ref{eq:z constriant}).

The remaining step now is to eliminate the $\hat{\psi}_i$ in (\ref{eq:sin}) and (\ref{eq:cos}) and meanwhile preserve the constraints implied therein, thus obtaining a set of relations equivalent to those in (\ref{eq:dp=Rv}). First, we have from (\ref{eq:sin})
\begin{equation}
	\begin{split}
		\sin \hat{\psi}_i=\left( \bar{\nu}_{i|x}\left( \hat{y}_{i}^{u}-\hat{y}_{i}^{v} \right) -\bar{\nu}_{i|y}\left( \hat{x}_{i}^{u}-\hat{x}_{i}^{v} \right) \right) /({\bar{\nu}_{i|x}}^2+{\bar{\nu}_{i|y}}^2) \\=\left( \bar{\nu}_{i|x}\left( \hat{y}_{i}^{u}-\hat{y}_{i}^{w} \right) -\bar{\nu}_{i|y}\left( \hat{x}_{i}^{u}-\hat{x}_{i}^{w} \right) \right) /({\bar{\nu}_{i|x}}^2+{\bar{\nu}_{i|y}}^2),\\ \forall u,v,w\in \mathcal{B}, u<v<w.
	\end{split}
\end{equation}
A similar relationship can be obtained from (\ref{eq:cos}). These two equalities appear in (\ref{eq:yaw}).

Moreover, (\ref{eq:sin}) and (\ref{eq:cos}) impose that $\sin ^2\hat{\psi }_i+\cos ^2\hat{\psi} _i=1$, i.e., 
$$
\left( \hat{x}_{i}^{u}-\hat{x}_{i}^{v} \right) ^2+\left( \hat{y}_{i}^{u}-\hat{y}_{i}^{v} \right) ^2={\bar{\nu}_{i|x}}^2+{\bar{\nu}_{i|y}}^2,
$$
which can be combined with (\ref{eq:dz}) to obtain a relation in a more compact form  
$$
\left( \hat{x}_{i}^{u}-\hat{x}_{i}^{v} \right) ^2+\left( \hat{y}_{i}^{u}-\hat{y}_{i}^{v} \right) ^2+\left( \hat{z}_{i}^{u}-\hat{z}_{i}^{v} \right) ^2={\bar{\nu}_{i|x}}^2+{\bar{\nu}_{i|y}}^2+{\bar{\nu}_{i|z}}^2,
$$
which is exactly the calibration constraint (\ref{eq:norm constraint}).

From the above derivation, we can state that given any sensor coordinates satisfying (\ref{eq:norm constraint}) (\ref{eq:z constriant}) and (\ref{eq:yaw}), we can an recover an $\{R_i\}$ satisfying (\ref{eq:dp=Rv}), i.e., satisfying (\ref{eq:equation}) according to (\ref{eq:p=Rv+t2dp=Rdv}), so that the aspect (b) of Proposition 1 holds. The aspect (a) is also naturally guaranteed by the above derivation.

\textbf{The 6-axis proprioception setup.} Eq. (\ref{eq:6 axis IMU}) is an instance of (\ref{eq:p=Rv+t2dp=Rdv})  with a fully observed $R_i$. The aspects (a) and (b) hold with no elaboration. \qed

\section{Finding a Equivalent Subset of Constraints in Problem 2}
\label{appendix II}
In this appendix, we follow an intuitive geometric idea to find an equivalent subset of the constraints of Problem 2.

\textbf{The distance-only setup.} It is well known that in a plane, a point can be uniquely determined by known distances to any 3 non-collinear known points. According to this, when $\left| \mathcal{B} _i \right|\geq 4$, supposing there exist 3 non-collinear distance sensors, we first choose 3 non-collinear sensors, donated as $p_i^0$, $p_i^1$ and $p_i^2$, and an equivalent subset of (\ref{eq:norm constraint}) for agent $i$ can be written as  
\begin{equation}
	\label{eq:norm constraint subset}
	\left\| p_{i}^{u}-p_{i}^{v} \right\| ^2=\left\| \bar{\nu}_{i}^{u}-\bar{\nu}_{i}^{v} \right\| ^2, \forall u\in \left\{ 0,1,2 \right\} ,v\in \mathcal{B} _i,u<v.
\end{equation}
The above set actually includes the minimum number of constraints to ensure equivalence with (\ref{eq:norm constraint}). 

When all the sensors are colinear, we only need to choose 2 sensors first and only maintain the calibration constraints between the other sensors and these 2 sensors. 

\textbf{The 4-axis proprioception setup.} When $\left| \mathcal{B} _i \right|\geq 5$ and 4 non-coplanar sensors exist, we follow a similar idea to the distance-only setup: considering that a point in space can be uniquely determined by known distances to any 4 non-coplanar points, we first specify 4 'anchor' non-coplanar distance sensors that satisfy both aspects of Proposition 1. In this way, only the calibration constraints (\ref{eq:norm constraint}) between other sensors and these anchor sensors are needed to preserved to guarantee the other sensors satisfy Proposition 1, while we can discard the proprioception constraints (\ref{eq:z constriant})-(\ref{eq:yaw}) between non-anchor sensors and the anchors. 

Therefore, we only need to preserve necessary constraints between the 4 anchors and the calibration constraints between non-anchor sensors and the anchors. Specially, we choose to only maintain all the linearly independent proprioception constraints among these 4 sensors \footnote{Although in some cases it is possible further to reduce the constraint set of these 4 anchor sensors, it requires an exhaustive case-by-case discussion} and the constraints of (\ref{eq:norm constraint}) between them and the other sensors. Let these 4 anchor sensors be $p_{i}^{0}$, $p_{i}^{1}$, $p_{i}^{2}$ and $p_{i}^{3}$, we can write this equivalent subset of the constraints as
\begin{equation}
	\left\| p_{i}^{u}-p_{i}^{v} \right\| ^2=\left\| \bar{\nu}_{i}^{u}-\bar{\nu}_{i}^{v} \right\| ^2, \forall u\in \left\{ 0,1,2,3 \right\} ,v\in \mathcal{B} _i,u<v
	\vspace{-0.2cm}
\end{equation}
$$		
		z_{i}^{u}-z_{i}^{v}=\left[ \begin{matrix}
			-\sin \tilde{\phi}_i,\	\cos \tilde{\phi}_i\sin \tilde{\theta}_i, \		\cos \tilde{\phi}_i\cos \tilde{\theta}_i\\
		\end{matrix} \right] \left( \bar{\nu}_{i}^{u}-\bar{\nu}_{j}^{v} \right)\hspace{-0.05cm}, 
$$
\vspace{-0.7cm}
\begin{equation}
	\label{eq:z subset}
	u,v,w\in \{0,1,2,3\} , u< v<w, \vspace{-0.5cm}
\end{equation}
\begin{center}
	\footnotesize$${\left( \bar{\nu}_{i|x}\left( y_{i}^{u}-y_{i}^{v} \right) -\bar{\nu}_{i|y}\left( x_{i}^{u}-x_{i}^{v} \right) \right)}={\left( \bar{\nu}_{i|x}\left( y_{i}^{v}-y_{i}^{w} \right) -\bar{\nu}_{i|y}\left( x_{i}^{v}-x_{i}^{w} \right) \right)}, $$
\end{center}
\vspace{-1cm}
\begin{center}
	\footnotesize$${\left( \bar{\nu}_{i|x}\left( x_{i}^{u}-x_{i}^{v} \right) +\bar{\nu}_{i|y}\left( y_{i}^{u}-y_{i}^{v} \right) \right)} = {\left( \bar{\nu}_{i|x}\left( x_{i}^{v}-x_{i}^{w} \right) +\bar{\nu}_{i|y}\left( y_{i}^{v}-y_{i}^{w} \right)\right)},$$
\end{center}
\vspace{-0.3cm}
\begin{equation}
	\label{eq:yaw subset}
	u,v,w\in \{0,1,2,3\} , u< v<w, \vspace{-0.2cm}
\end{equation}
where we do not explicitly remove the linearly dependent constraints, as it is trivial to identify them.

When all the sensors are coplanar, we just need first to characterize 3 anchor sensors and redo the above derivation to obtain a set of constraints analogous to (\ref{eq:norm constraint subset}) and (\ref{eq:yaw subset}).
 
\textbf{The 6-axis proprioception setup.} This setup is very similar to a distance-only setup, so we directly give the result when  $\left| \mathcal{B} _i \right|\geq 5$ as follows:
\begin{equation}
	\label{eq:6 axis IMU subset}
	p_{i}^{u}-p_{i}^{v}=\hat{R}_i\left( \bar{\nu}_{i}^{u}-\bar{\nu}_{i}^{v} \right), \ \forall u\in \left\{ 0,1,2,3 \right\} ,v\in \mathcal{B} _i,u<v.
\end{equation}

\section{Derivation of Results about ESDP}
\label{appendix III}
\subsection{Proof of Proposition 3}
\emph{Proof:} If a certain sensor $(i,u)$ is not uniquely localizable, then there are at least two different solutions of $p_i^u$ in the solution set. On the other hand, Theorem 4.2 in \cite{wang2008further} says that if   $X_i^u=(p_i^u)^Tp_i^u$, then $p_i^u$ is invariant in the solution set. So we must have  $X_i^u>(p_i^u)^Tp_i^u$ considering constraint (\ref{eq:ESDR}) (which implies that the relaxation is not tight). \qed
\subsection{Derivation of Eq. (\ref{eq:X>p2}) and Eq. (\ref{eq:edge constraint})}
Constraint (\ref{eq:ESDR}) can be rewritten in a similar form as (\ref{eq:X>ptp}):
$$
\left[ \begin{matrix}
	X_{ii}^{uu}&		X_{ij}^{uv}\\
	X_{ij}^{uv}&		X_{jj}^{vv}\\
\end{matrix} \right] \succeq \left[ \begin{matrix}
	p_{i}^{u}&		p_{j}^{v}\\
\end{matrix} \right] ^T\left[ \begin{matrix}
	p_{i}^{u}&		p_{j}^{v}\\
\end{matrix} \right], 
$$
i.e.,
$$
\left[ \begin{matrix}
		X_{ii}^{uu}-\left( p_{i}^{u} \right) ^Tp_{i}^{u}&		X_{ij}^{uv}-\left( p_{i}^{u} \right) ^Tp_{j}^{v}\\
	X_{ij}^{uv}-\left( p_{j}^{v} \right) ^Tp_{i}^{u}&		X_{jj}^{vv}-\left( p_{j}^{v} \right) ^Tp_{j}^{v}\\
\end{matrix} \right] \succeq 0,
$$
of which all of the leading principal minors in the matrix are nonnegative, i.e.,
$$
	X_{ii}^{uu}-\left( p_{i}^{u} \right) ^Tp_{i}^{u}\geq 0,
$$
\begin{equation}
	\label{eq:ESDR sub}
	\begin{split}
		\left| \begin{matrix}
			X_{ii}^{uu}-\left( p_{i}^{u} \right) ^Tp_{i}^{u}&		X_{ij}^{uv}-\left( p_{i}^{u} \right) ^Tp_{j}^{v}\\
			X_{ij}^{uv}-\left( p_{j}^{v} \right) ^Tp_{i}^{u}&		X_{jj}^{vv}-\left( p_{j}^{v} \right) ^Tp_{j}^{v}\\
		\end{matrix} \right|= \\ \left( X_{ii}^{uu}-\left( p_{i}^{u} \right) ^Tp_{i}^{u} \right) \left( X_{jj}^{vv}-\left( p_{j}^{v} \right) ^Tp_{j}^{v} \right) - \\ \left( X_{ij}^{uv}-\left( p_{i}^{u} \right) ^Tp_{j}^{v} \right) ^2\geq 0.
		\\
	\end{split}
\end{equation}
Fixing the components corresponding to the sensor on agent $j$, i.e., $p_{j}^{v}$ and $X_{jj}^{vv}$,  in (\ref{eq:ESDR sub}),  we can obtain the constraint (\ref{eq:edge constraint}) by organizing.

\section{Proof of Lemmas in Section \ref{sec:6C}}
\label{appendix IV}
\subsection{Proof of Lemma 3}
\emph{Proof:} \textbf{Boundness of a single element in Problem 5}. Assume $(i,u)$ is a sensor that is directly connected to a prior-known sequence-bounded sensor $(a,v)$ guaranteed to exist by Assumption 2. Then we have 
\begin{equation}
	\begin{split}
		\kappa((U_i^u-U_a^v)^T(V_i^u-V_a^v)-\tilde{q}_{ia}^{uv})^2 + \gamma \left\| U_i^{u}-V_i^u \right\| _{F}^{2} \\ 
		\leq 
		F\left( \bar{U},\bar{V} \right), 
	\end{split}
	\label{eq:residual}
\end{equation}
where $(U_{i}^{u}$, $V_{i}^{u})$ and $(U_{a}^{v}$, $V_{a}^{v})$ are the corresponding surrogates/column of sensor $(i,u)$ and $(a,v)$ in variables $U$ and $V$, which are in the level set  $\mathcal{H}_F((\bar{U},\bar{V}))$, where $F(\cdot)$ is the objective function in Problem 5. $\kappa$ in (\ref{eq:residual}) is $\frac{1}{ (\sigma _{ia}^{uv})^2 }$ in (\ref{eq:problem 2}) if the sensor $(i,u)$ and the sensor $(a,v)$ are on the different agents, or is $\lambda$ in (\ref{eq:cost 2}) if they are on the same agent. According to Assumption 2, $\{ \left( p_{a}^{v} \right) ^{\left( k \right)} \} $, i.e.,   $\{ (( U_{a}^{v} ) ^{\left( k \right)}, ( V_{a}^{v} ) ^{\left( k \right)}) \} $, is bounded. Therefore,  $\{ (( U_{i}^{u} ) ^{\left( k \right)}, ( V_{i}^{u} ) ^{\left( k \right)}) \} $ must be bounded, otherwise (\ref{eq:residual}) does not hold, so that variables corresponding to any sensor connected (on the same agent as or have a measurement with) the sensor $(i,u)$ can be proved to be bounded in the same way.

\textbf{Boundness of a single element in Problem 4.} We first have (\ref{eq:X>p2}) so that 
\begin{equation}
	\label{eq:residual2}
	\begin{split}
		\sqrt{\kappa}\left( \left( p_{i}^{v} \right) ^Tp_{i}^{u}-2\left( p_{a}^{v} \right) ^Tp_{i}^{u}+X_{aa}^{vv}-\tilde{q}_{ia}^{uv} \right) \\ \leq \sqrt{\kappa}\left( X_{ii}^{uu}-2\left( p_{a}^{v} \right) ^Tp_{i}^{u}+X_{aa}^{vv}-\tilde{q}_{ia}^{uv} \right) \leq \sqrt{G( \bar{Z} )},
	\end{split}
\end{equation}
where the variables are in the level set $\mathcal{H}_G(\bar{Z})$ as previous and $G( \bar{Z} )$ is the objective function in Problem 4. According to Assumption 2, $\{ \left( p_{a}^{v} \right) ^{\left( k \right)} \} $ is bounded, so that $p_{i}^{u}$ must be bounded indicated by (\ref{eq:residual2}). Then, we immediately have $X_{ii}^{uu}\leq \sqrt{{{G( \bar{Z} )}/{\kappa}}}+2\left( p_{a}^{v} \right) ^Tp_{i}^{u}-X_{aa}^{vv}+\tilde{q}_{ia}^{uv}$ from (\ref{eq:residual2}), which indicates that $X_{ii}^{uu}$ is also bounded.

Let $(j,w)$ be a sensor directly connected with sensor $(i,u)$. We have 
\begin{equation}
	\label{eq:XX>XX}
	X_{ii}^{uu}X_{jj}^{ww}\geq \left( X_{ij}^{uw} \right) ^2
\end{equation} 
so that 
\begin{equation}
	\label{eq:residual3}
	\begin{split}
		\sqrt{\kappa}\left( X_{ii}^{uu}-2\sqrt{X_{ii}^{uu}X_{jj}^{ww}}+X_{jj}^{ww}-\tilde{q}_{ij}^{uw} \right) \\ \leq \sqrt{\kappa}\left( X_{ii}^{uu}-2X_{ij}^{uw}+X_{jj}^{ww}-\tilde{q}_{ij}^{uw} \right) \leq \sqrt{G\left( \bar{Z} \right)}.
	\end{split}
\end{equation}
Since $X_{ii}^{uu}$ is bounded, $X_{jj}^{ww}$ is bounded from (\ref{eq:residual3}). According to (\ref{eq:X>p2}) and (\ref{eq:XX>XX}), $p_j^w$ and $X_{ij}^{uw}$ are also bounded. The above proof can be applied in the same way between different sensors on the same agent.

\textbf{Boundness of all elements.} Since Assumption 2 says that any sensor is directly or indirectly connected to a prior-known bounded-sequence sensor, then we can traverse any path in $\mathcal{G}$ from the bounded-sequence sensor to that sensor and apply the above proof in succession between neighboring sensors. In this way, the boundedness of the variable corresponding to any sensor can be obtained. \qed

\subsection{Proof of Lemma 4}
\emph{Proof:} We follow a similar way to Proposition 3.9 in \cite{bertsekas2015parallel} to prove this lemma. Let $\boldsymbol{x}^*=\left( x_{1}^{*},...,x_{p}^{*} \right)$ be a Nash equilibrium of (\ref{eq:opt}) such that
$$
	f\left( x_{1}^{*},...,x_{q}^{*},...x_{p}^{*} \right) \leq f\left( x_{1}^{*},...,x_q,...x_{p}^{*} \right), \ \forall x_{q} \in \mathcal{X} _q
$$
holds for all $q=1,...,p$ according to Defination 3, i.e., $x_{q}^{*}$ is a global minimizer of $f\left( x_{1}^{*},...,x_q,...x_{p}^{*} \right)$. So we have 
\begin{equation}
\label{eq:optimality}
\nabla _qf\left( \boldsymbol{x}^* \right) ^Td=\nabla f_q^*( {x}_{q}^{*}) ^Td\geq 0, \  \forall d\in \mathcal{T}^* _{\Omega _q}( x_{q}^{*}) 
\end{equation}
where $
f_q^*\left( x_q \right) \coloneqq f( x_q;\hat{\boldsymbol{x}}^*_{\left[ p \right] /\left\{ q \right\}}) 
$ is the block objective function of $x_q$ with remaining blocks fixed as $\boldsymbol{x}^*$ and $\mathcal{T}^* _{\Omega _q}( x_{q}^{*}) $ is the tangent cone \cite[Defination 12.2]{nocedal1999numerical} of $f_q^*(\cdot)$ at $x_q^*$.

Since the block division makes $\mathcal{X}$ a Cartesian product over each block, it follows that (\ref{eq:optimality}) holds for $d$ in all tangent cones of \emph{arbitrarily} fixed remaining blocks at $x_{q}^{*}$. So by adding (\ref{eq:optimality}) over all $q$, we have 
\begin{equation}
	\nabla f\left( \boldsymbol{x}^* \right) ^Td\geq 0, \ \forall d\in \mathcal{T} _{\Omega}( \boldsymbol{x}^* ), 
\end{equation}
where $\mathcal{T} _{\Omega}( \boldsymbol{x}^* )$ is the tangent cone of the full object function $f(\cdot)$. Therefore, $\boldsymbol{x}^*$ is a local solution of (\ref{eq:opt}) according to Theorem 12.3 in \cite{nocedal1999numerical} and under some constraint qualification it can be rewritten in the form of the KKT condition. \qed

\section{Various $r$ in BM-BCD}
\label{appendix V}
\begin{figure} \centering 
	\setcounter{subfigure}{0} 
	\hspace{-0.3cm}
	\subfigure[]{
		\raisebox{-0.03cm}
		{\includegraphics[width=0.46\columnwidth]{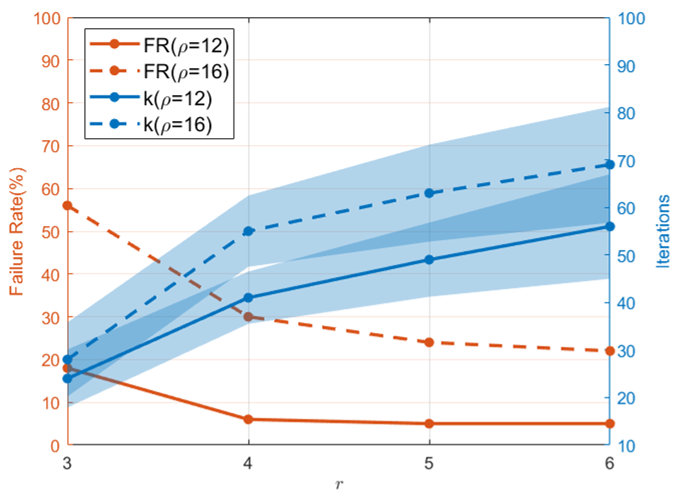}
			\label{fig:r-FR}}
	}      
	\subfigure[]{  
		{\includegraphics[width=0.41\columnwidth]{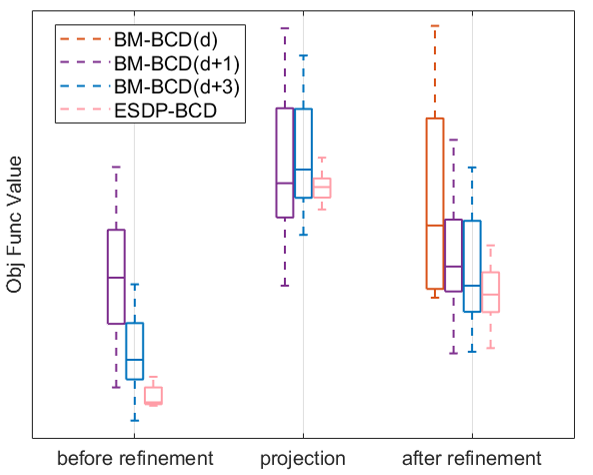}
			\label{fig:obj}}     
	} \vspace{-0.2cm} \\  
	\captionsetup{labelformat=simple}
	\captionsetup{font=footnotesize}  
	\caption{\textbf{BM-BCD with various $r$ and under harsh initializations.}  Statistics are computed under 100 times of data generation. (a) The failure rate and iterations needed for each $r$. (b) The resulting objective function value of different stages in the solution process, where 'before refinement', 'projection', and 'after refinement' respectively indicate the objective function values of Problem 4 or Problem 5, the objective function values of Problem 2 at estimates extracted from Problem 5 (i.e., $Q^TU$) or Problem 4 (the (1,2) block of $Z$), and the objective function values of Problem 2 after refinement.}   
	\label{fig:r}
\end{figure}
In this appendix, we study how the parameter $r$ in BM-BCD affects its performance. We conduct experiments in the 3D \textbf{pyramid} problem using the same setup as Section \ref{sec:7}. We use poor initializations to trigger failure and record the failure rate (FR), defined as the rate RMSE exceeds 60 cm.

\textbf{Results.} From Fig. \ref{fig:r-FR}, we can see that the failure rate hopefully decreases as the rank $r$ becomes larger.  The story behind this observation, which can be revealed in Fig. \ref{fig:obj}, is that a larger $r$, such as that in BM-BCD($d$+3), can lead to a more consistent solutions under poor initializations, which can be captured by the low variance of the corresponding objective function values. This implies that shallow local minima of the optimization model are seemingly reduced as $r$ increases. Thus, a higher $r$ provides more consistent (although not always better) initializations for the refinement stage after projecting the high-rank solution onto the realization space than those of lower $r$. However, we can only empirically claim the robustness from higher-rank spaces so far, although there are remarkable results about the strong relationship between local minima of NLP problems from BM factorization and the global minimizer of a class of SDP \cite{boumal2015riemannian}.

\begin{figure} \centering 
	\vspace{-0.5cm}
	\setcounter{subfigure}{0} 
	\hspace{-0.3cm}
	\subfigure{
		{\includegraphics[width=0.468\columnwidth]{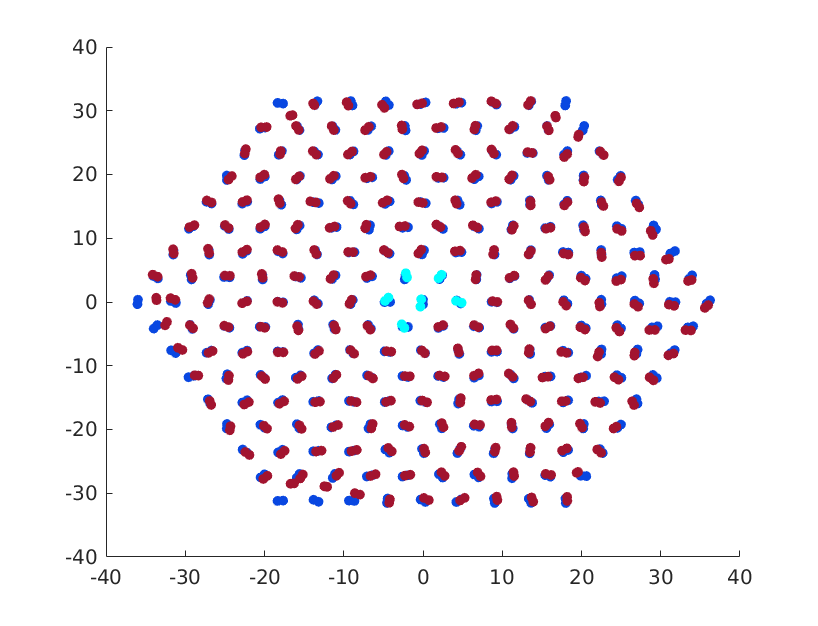}
			\label{fig:hexagon1}}
	} 
	\hspace{-0.3cm}     
	\subfigure{  
		{\includegraphics[width=0.47\columnwidth]{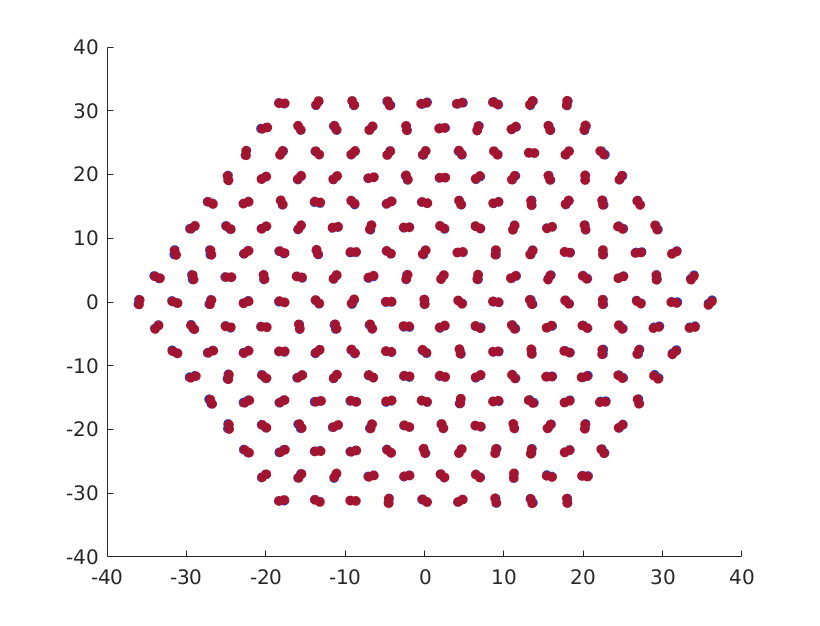}
			\label{fig:hexagon2}}     
	}  \\  
	\captionsetup{labelformat=simple}
	\captionsetup{font=footnotesize}
	\vspace{-0.3cm}  
	\caption{ \textbf{Visualization of the problem \textnormal{hexagon} and results.} Results before (left) and after (right) refinement of ESDP-BCD in \textbf{hexagon}, where the graphic elements have the same meaning as in Fig. \ref{fig:combination}.} 
	\label{fig:hexagon}
\end{figure}

\begin{figure} \centering 
	\vspace{-0.3cm}
	\setcounter{subfigure}{0} 
	\hspace{-0.3cm}
	\subfigure{
		{\includegraphics[width=0.5\columnwidth]{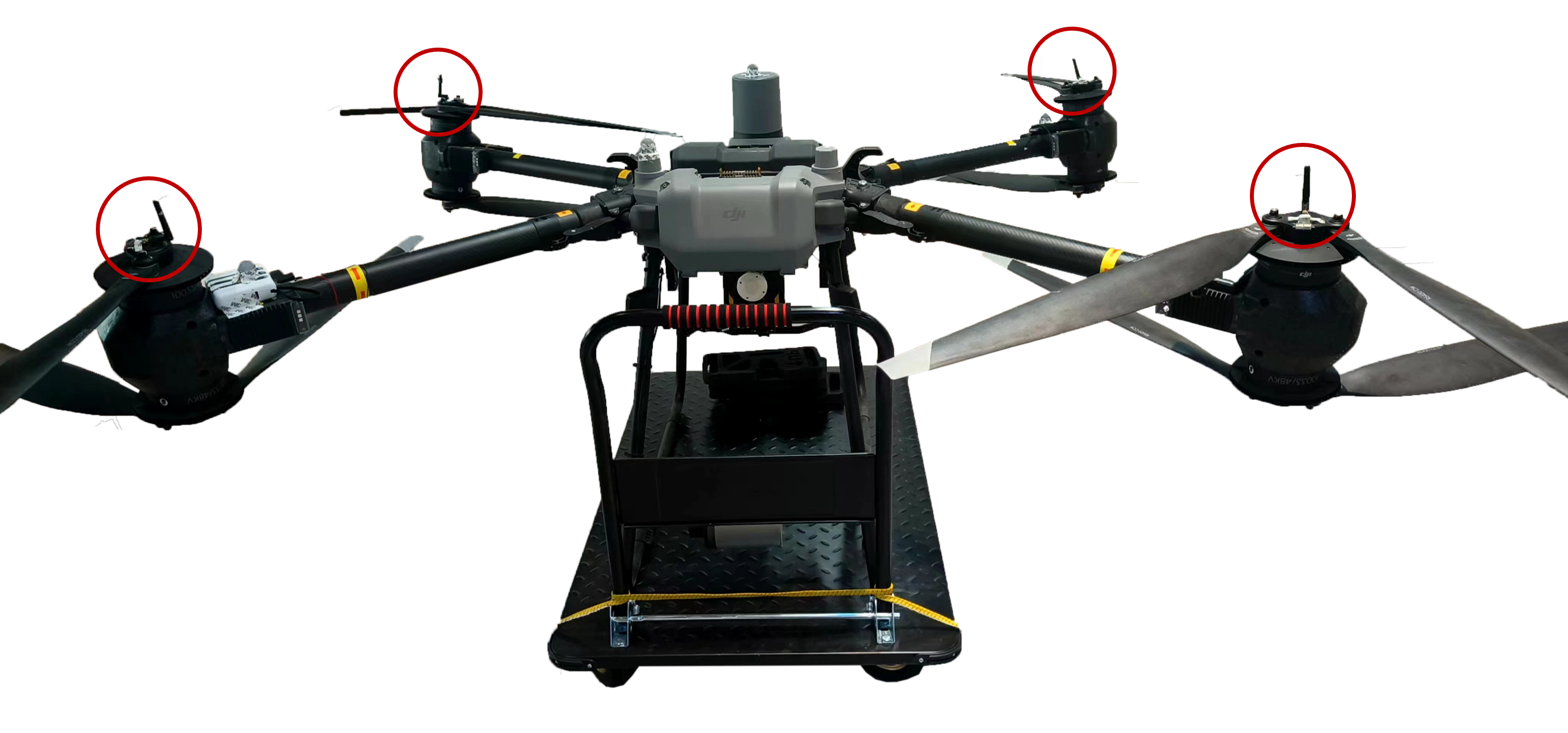}
			\label{fig:large}}
	} 
	\hspace{-0.3cm}     
	\subfigure{  
		{\includegraphics[width=0.35\columnwidth]{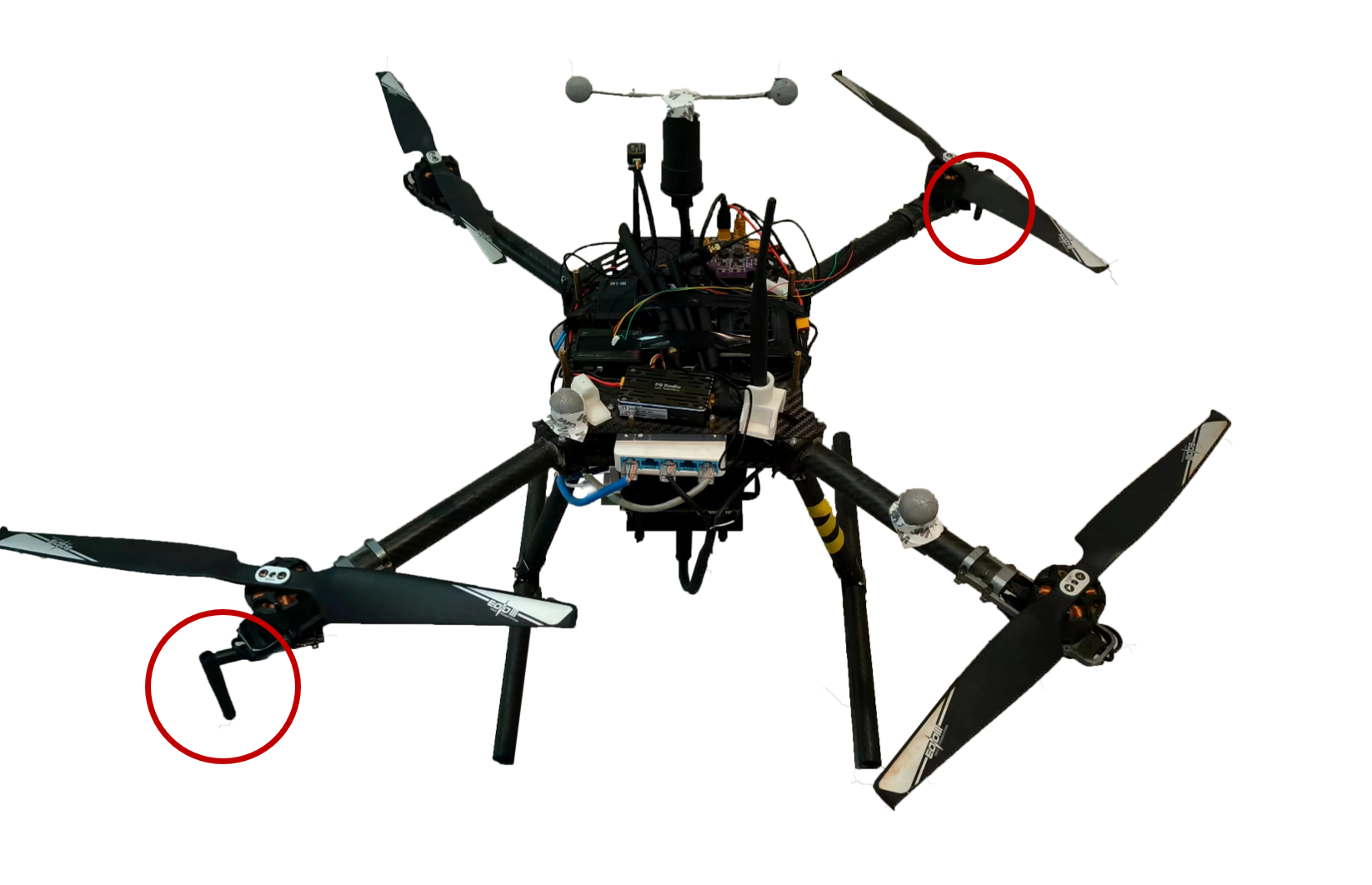}
			\label{fig:small}}     
	}  \\  
	\captionsetup{labelformat=simple}
	\captionsetup{font=footnotesize}
	\vspace{-0.3cm}  
	\caption{ \textbf{Illustration of the used hardware platforms.} The mothership (left) and parasite (right) used in our experiment. The dark red circles highlights the equipped UWB sensors.} 
	\vspace{-0.2cm}
	\label{fig:hardware}
\end{figure}

\begin{figure} \centering 
	\vspace{-0.5cm}
	\setcounter{subfigure}{0} 
	\hspace{-0.3cm}
	\subfigure{
		{\includegraphics[width=0.7\columnwidth]{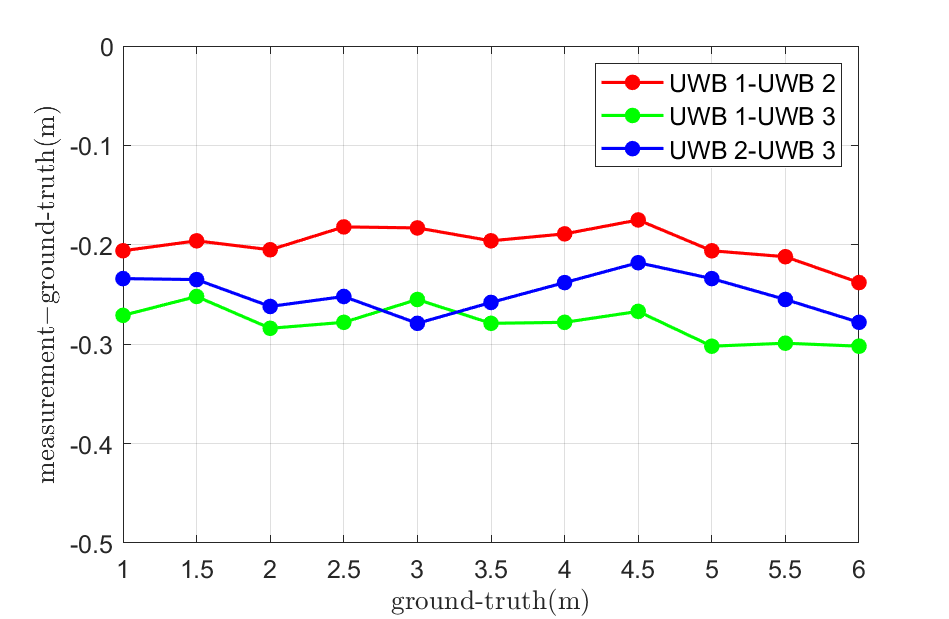}
	} }
	\captionsetup{labelformat=simple}
	\captionsetup{font=footnotesize}
	\vspace{-0.3cm}  
	\caption{ \textbf{Mean biases between different pairs among 3 UWBs.} The y-axis is the mean bias, which is calculated as the average of (measurement$-$ground-truth) of the collected data.} 
	\label{fig:calibration}
\end{figure}

\section{Detailed Setup and Method of the Real-world Experiment in Section \ref{sec:7C}}
\label{appendix VI}
\subsection{Hardware Platform}
We use a DJI FlyCart 30 quadcopter to serve as our mothership, as shown in Fig. \ref{fig:large}. One of the platforms serving as the parasite is shown in Fig. \ref{fig:small}. Nooploop LTP-AS2 UWB module is used to measure distances. We use a motion capture (mocap) system to record the ground-truth trajectories, and the origin of the body coordinate system of each agent is determined by the placement of the mocap markers on it. The calibrated distance sensors on the mothership is $\bar{\nu}_{0}^{0}=\left[ 0.725;-0.897;-0.06 \right]$m, $\bar{\nu}_{0}^{1}=\left[ -0.691;-0.897;-0.06 \right]$m, $\bar{\nu}_{0}^{2}=\left[ -0.691;0.837;0.04 \right]$m, and $\bar{\nu}_{0}^{3}=\left[ 0.725;0.837;0.04 \right]$m. The calibrated coordinates of the distance sensors on each parasite vary slightly beacuse the layout of mocap markers is different on each parasite, but the length of the baselines are all around $0.7$m.

\subsection{Data Processing}
As observed in \cite{fishberg2022multi,fishberg2023murp}, the Nooploop UWB modules generate data in distributions with non-zero means (called \emph{bias}). To narrow the gap from simulation to the reality, we post-process the reading of UWBs to approximately satisfy the zero-mean assumption used in our formulation (see (\ref{eq:measurement model0})).

Before collecting the dataset on which we run the experiment, we first collect a calibration dataset to decide a strategy for processing measurements. We find that the biases of the measurements are relatively stable when the operating conditions do not vary much, while varying among different pairs of sensors, as shown in Fig. \ref{fig:calibration}. However, these biases can be significantly different (e.g., with a difference greater than $10$ cm) when there is occlusion or when a large attitude difference between the two UWB sensors, as previously observed in \cite{fishberg2023murp}. When collecting our dataset, we artificially control the relative altitude between UWBs so as not to be too large and avoid occlusions between them. We believe that with further advancement of distance sensors and a more elaborate system design, such manual operation can be avoided.

Once the above pattern is found, we employ a simple calibration scheme: we use the calibration dataset to compute an average bias for each pair of UWBs where measurements exist, and then subtract this bias from measurements generated by the corresponding UWB pair when running the experiment. 

\subsection{Notes on the Results}
	We note that we use a near perfect relative estimate (from the mocap system) to initialize the two methods at moment $t=0$ so that the local search method does not fall into a shallow local minima at the beginning, making it easier to exhibit the gradually increasing relative error $\eta$. When giving a good initialization at beginning, we find that if a higher frequency (e.g., 50Hz) of the measurement is available (we simulate it using high-frequency mocap data), the estimation drift of the local search method is mitigated if we run it at correspondingly high frequencies. This is because when the frequency is high, the initialization, i.e., the estimation of last frame, of the optimization corresponding to each frame of measurement can be viewed as being better-maintained than that of lower frequency of measurements. However, high-frequency online estimation greatly enhances the requirements for onboard computation as well as communication. We therefore believe that the convex relaxation approach has a unique interest in maintaining time-consistent estimates even when the local search frequency is not that high.

\end{document}